\documentclass{article}

\usepackage{microtype}
\usepackage{graphicx}
\usepackage{booktabs} 
\usepackage[numbers]{natbib} 
\usepackage{graphicx,subfig}
\usepackage[T1]{fontenc}
\usepackage{changepage}
\usepackage{tcolorbox}
\usepackage{hyperref}
\usepackage{datatool}
\usepackage{graphicx}
\usepackage{multirow}
\usepackage{thmtools}
\newcommand{\state}{s}

\definecolor{black}{rgb}		{0.0, 0.0, 0.0}
\definecolor{white}{rgb}		{1.0, 1.0, 1.0}
\definecolor{yellow}{rgb}		{1.0, 1.0, 0.8}
\definecolor{red}{rgb}			{0.6, 0.0, 0.2}
\definecolor{blue}{rgb}		{0.0, 0.2, 0.5}
\definecolor{green}{rgb}		{0.6, 0.8, 0.8}
\definecolor{dark_green}{RGB} {0, 140, 0}
\definecolor{gold}{rgb}		{0.6, 0.4, 0.1}
\definecolor{grey}{RGB}{0,0,0}
\definecolor{Gray}{gray}{0.8}
\definecolor{MediumGray}{gray}{0.9}
\definecolor{LightGray}{gray}{0.98}
\definecolor{LightCyan}{rgb}{0.88,1,1}
\definecolor{purple}{RGB}{128,0,128}
\definecolor{sl_blue}{RGB}{47, 60, 105}
\definecolor{orange}{RGB}{255,165,0}
\definecolor{Gray}{gray}{0.85}

\usepackage{etoolbox}
\usepackage{amsfonts}
\usepackage{amsmath}
\usepackage{amsthm}
\usepackage{xspace}
\usepackage{enumerate}
\usepackage{bm}

\providecommand{\aA}[1]{\argsA{\AC}{#1}}
\providecommand{\argsA}[2]{ {#1}_{#2} } 
\newcommand{\mop}{mixture of policies\xspace}
\DeclareMathOperator*{\argmax}{argmax}

\newcommand{\gammapbrs}{\gamma_{\phi}}
\newcommand{\exploitabilityi}{\mathcal{G}_{E,i}}
\newcommand{\exploitabilityj}{\mathcal{G}_{E,j}}

\newcommand{\E}{\mathbb{E}}

\newcommand{\M}{\mathcal{M}}

\newcommand{\states}{\mathcal{S}}

\newcommand{\actions}{\mathcal{A}}
\newcommand{\action}{a}

\newcommand{\rewards}{\mathcal{R}}
\newcommand{\transition}{\mathcal{P}}
\newcommand{\observations}{\Omega}
\newcommand{\observationf}{\mathcal{O}}

\newcommand{\red}{Red\xspace}
\newcommand{\blue}{Blue\xspace}
\newcommand{\eg}{e.g., }  
\newcommand{\ie}{i.e., }  

\newcommand{\commentout}[1]{}
\newcommand{\realnumber}{\mathbb{R}}

\newcommand{\ADO}{{ADO}\xspace}

\global\long\def\JointPolicy#1#2{\langle #1, #2 \rangle}
\newtheorem{definitionsec}{Definition}[section] 
\newtheorem{theoremsec}{Theorem}
\providecommand{\say}[1]{``#1''} 
\providecommand{\gets}{$\leftarrow$}
\providecommand{\ExtensiveFormGame}{\mathcal{M}}
\providecommand{\NormalFormGame}{\mathcal{N}}
\providecommand{\Exploitability}{\mathcal{G}_{E}}
\providecommand{\AC}{\pi}                 
\providecommand{\funcName}[1]{\textsc{#1}}
\global\long\def\mA#1{\argsA{\mu}{#1}}
\global\long\def\G#1#2{\argsA{\mathcal{G}}{#1}(#2)}

\global\long\def\JointPolicy#1#2{\langle #1, #2 \rangle}
\global\long\def\O#1#2{\argsA{O}{#1}(#2)}

\usepackage[framemethod=tikz]{mdframed}
\newmdenv[innerlinewidth=0.5pt, roundcorner=4pt,linecolor=red,innerleftmargin=6pt,
innerrightmargin=6pt,innertopmargin=6pt,innerbottommargin=6pt]{mybox}
\newcommand{\MRO}{MRO\xspace}
\newcommand{\mro}{MRO\xspace}

\theoremstyle{definition}
\setcitestyle{citesep={;}}

\usepackage[accepted]{icml2025}
\usepackage{amsmath}
\usepackage{amssymb}
\usepackage{mathtools}
\usepackage{amsthm}

\theoremstyle{plain}
\newtheorem{theorem}{Theorem}[section]

\theoremstyle{definition}

\newtheorem{assumption}[theorem]{Assumption}
\theoremstyle{remark}

\usepackage{fancyhdr}
\fancypagestyle{plain}{%
\fancyhead{} 
\fancyfoot{} 
\fancyfoot[CE,CO]{\thepage  \\ \small Unpublished Work. Copyright 2025 BAE Systems. All Rights Reserved.}}

\renewcommand{\cite}[2][]{\citeauthor{#2} (\citeyear[#1]{#2})\xspace}

\icmltitlerunning{An Empirical Game-Theoretic Analysis of Autonomous Cyber-Defence Agents}

\begin{document}


\twocolumn

\twocolumn[
\icmltitle{An Empirical Game-Theoretic Analysis of Autonomous Cyber-Defence Agents}



\icmlsetsymbol{equal}{*}

\begin{icmlauthorlist}
\icmlauthor{Gregory Palmer}{BAE}
\icmlauthor{Luke Swaby}{BAE}
\icmlauthor{Daniel J.B. Harrold}{BAE}
\icmlauthor{Matthew Stewart}{BAE}
\icmlauthor{Alex Hiles}{BAE}
\icmlauthor{Chris Willis}{BAE}
\icmlauthor{Ian Miles}{FNC}
\icmlauthor{Sara Farmer}{DSTL}
\end{icmlauthorlist}

\icmlaffiliation{BAE}{BAE Systems Applied Intelligence Labs, United Kingdom (UK)}
\icmlaffiliation{FNC}{Frazer-Nash Consultancy Limited, UK}
\icmlaffiliation{DSTL}{Defence Science and Technology Laboratory (Dstl), UK}

\icmlcorrespondingauthor{Gregory Palmer}{gregory.palmer@baesystems.com}

\icmlkeywords{Reinforcement Learning, Autonomous Resilient Cyber Defence}

\vskip 0.3in
]



\printAffiliationsAndNotice{}  

\begin{abstract}
The recent rise in increasingly 
sophisticated cyber-attacks raises the need 
for robust and resilient autonomous cyber-defence (ACD) agents.
Given the variety of cyber-attack tactics, 
techniques and procedures (TTPs) employed, learning approaches 
that can return \emph{generalisable} policies are desirable.
Meanwhile, the assurance of ACD agents  
remains an open challenge.
We address both challenges via an \emph{empirical 
game-theoretic analysis} of deep reinforcement learning (DRL) 
approaches for ACD using the principled \emph{double oracle} (DO) algorithm.  
This algorithm relies on adversaries iteratively learning  
(approximate) best responses  
against each others' policies; a 
computationally expensive endeavour for autonomous cyber operations agents.  
In this work we introduce and evaluate a theoretically-sound,
potential-based reward shaping approach to expedite
this process.
In addition, given the increasing number of open-source ACD-DRL approaches,
we extend the DO formulation to allow for \emph{multiple response oracles} (MRO),
providing a framework for a holistic evaluation of 
ACD approaches. 
\end{abstract}

\section{Introduction} \label{Introduction}

Deep reinforcement learning (DRL) has emerged as a promising approach
for training autonomous cyber-defence (ACD) agents that are capable of 
continuously investigating and neutralising cyber-threats at machine speed.
However, to learn robust and resilient ACD policies, DRL agents must  
overcome three open problem areas: 
i.)~Vast, dynamic, high-dimensional state-spaces; 
ii.)~Large, combinatorial action
spaces, and; 
iii.)~\emph{Adversarial learning} 
against a non-stationary opponent~\citep{palmer2023deep}. 

While the autonomous cyber-operations (ACO) community has taken 
steps towards solving the first two challenges~\citep{10.1145/3689933.3690835,tran2022cascaded}, 
the third challenge has received comparatively little attention.  
ACD agents are often benchmarked against a stationary set of autonomous cyber-attacking (ACA) agents.
For example, \emph{The Technical Cooperation Program} (TTCP)'s annual 
CybORG\footnote{Cyber Operations Research Gym~\citep{cyborg_acd_2021}.} CAGE Challenges (CCs) rank submissions 
based on their performance against the same rules-based 
ACA agents that are used during training~\citep{cage_challenge_2_announcement}.
This formulation is concerning from a generalisation standpoint; 
it encourages solutions that over-fit 
to known adversaries.
In real-world scenarios, by contrast, the stationarity of an opponent is a strong assumption. 
Furthermore, ACD agents will often be unable to perform online policy updates once deployed, effectively limiting them to a fixed-policy regime. 
Therefore, a framework is required for learning and assuring generalisable ACD policies \emph{prior to deployment}. 
In this work we advocate using variations of the principled \emph{double oracle} (DO) algorithm~\citep{mcmahan2003planning}
to obtain ACD agents that can generalise across ACA agents and 
provide assurance regarding their performance against a \emph{resource bounded worst case opponent}.\footnote{Bounded by computational resources~\citep{oliehoek2018beyond}.}

Our contributions can be summarised as follows:

\noindent{\textbf{i.)}} 
The DO algorithm
offers flexibility with respect to selecting algorithms for learning responses.\footnote{Referred to as \emph{policy space response oracles} (PSRO) when using DRL oracles~\citep{lanctot2017unified}.}
The availability of cyber-defence gyms 
has resulted in numerous open source (OS) ACD-DRL approaches,
which have frequently been infused with domain knowledge 
to simplify learning tasks, \eg wrappers for pre-processing observations and buffers for storing 
critical information~\citep{vyas2023automated}.
This raises questions such as: 
\say{Which approach should ACD practitioners select?} and
\say{Are there benefits in using more than one approach?}
To answer these questions, we extend the DO algorithm to allow for the computation
of responses using multiple approaches, 
giving rise to the \emph{multiple response oracles} algorithm (\MRO). 
We also add a theoretical evaluation of \MRO,
showing that the convergence guarantees from the underlying
DO algorithm remain intact.
%
%

\noindent{\textbf{ii.)}} 
While the DO algorithm offers the benefit of theoretical convergence 
guarantees, it suffers from lengthy wall-times 
due to the requirement that both adversaries iteratively compute 
(approximate) best responses (ABRs) against each
others' latest policies~\citep{li2023solving}. 
This challenge is amplified for autonomous cyber-operations (ACO), 
which confront learning agents with high-dimensional 
observation and action spaces.
Therefore, we are interested in extensions that facilitate the learning of ABRs
and reduce the number of response iterations for achieving convergence.
Here, we propose leveraging the lessons learned by previous 
responses by using their \emph{value functions}
for \emph{potential-based reward shaping} (VF-PBRS)~\citep{ng1999policy}.  
In addition, instead of computing responses from scratch, we initialise 
policies from pre-trained models (PTMs)~\citep{li2023solving}.

\noindent{\textbf{iii.)}} 
Using the MRO algorithm, 
we provide empirical game-theoretic analyses of 
state-of-the-art ACD-DRL approaches 
for two cyber-defence gym environments: 
CCs 2 and 4~\citep{cage_challenge_2_announcement,cage_challenge_4_announcement}.
We find that the ACD agents obtained via 
adversarial learning 
are robust towards learning attackers, 
which struggle to find new successful 
tactics, techniques and procedures (TTPs).
%
%
%
In addition, we provide empirical evidence that VF-PBRS oracles can converge upon significantly 
stronger ACD policies compared to vanilla approaches.
Finally, while pre-trained models allow for shorter response iterations,
in CC2 we find that additional \say{full} approximate best response iterations are required 
to achieve convergence,
with cyber-attacking responses using random neural network initialisations. 
%

%
%
%
%
%
%

\section{Related Work} \label{sec:related_work}

Despite ACD being, by definition, an adversarial game between defending and attacking agents,
there exists limited work on the topic of adversarial learning. 
Evaluations typically feature a stationary set of opponents~\citep{10.1145/3605764.3623986,nguyen2020multiple,o2024multi,tran2022cascaded}.
A possible explanation for this is that current cyber-defence gyms do not
support adversarial learning. 
For example, while the CybORG CCs support both the training of ACD and ACA agents,
modifications are necessary to enable learning against a DRL opponent. 
An exception here is work conducted by~\cite{shashkov2023adversarial}, who adapt and compare DRL, 
evolutionary strategies, and Monte Carlo tree search methods within CyberBattleSim~\citep{cyberbattlesim}. 
To the best of our knowledge, we are the first to conduct a holistic empirical game-theoretic analysis
of ACD-DRL approaches within topical cyber-defence gyms.

The topic of reward shaping has been gathering attention from the ACD community~\citep{lopes2023age,miles2024reinforcement}.
For instance, to study the sensitivity of DRL to the magnitude of received penalties~\citep{10.1145/3605764.3623916}.
In contrast to existing work, we seek a principled reward shaping function that will
preserve the DO algorithm's convergence guarantees.
Here, we identify the principled PBRS as a suitable approach,
and propose a novel formulation that utilises VFs from previous ABR iterations 
as a \emph{potential function}.
Finally, \cite{li2023solving} highlight the benefits of using DO training with PTMs for solving large-scale pursuit-evasion games.  
We believe we are the first to make use of this approach within the context of autonomous cyber-defence. 

\section{Background} \label{sec:background}

\subsection{Partially Observable Markov Games} \label{sec:background:pomg}

Our ACD and ACA agents are situated within
partially observable Markov games (POMGs). 
A POMG $\M$
is defined as a tuple $(n, \states, \observations, \observationf, \actions, \transition, \rewards, \gamma)$, 
consisting of: a finite state space $\states$;
a joint action space $(\actions_1 \times ... \times \actions_n)$ for each state
$\state \in \states$, with $\actions_p$ being the set of actions available 
to player $p$;
a state transition function $\transition : \states_t \times \actions_1 \times ... \times \actions_n \times \mathcal{\states}_{t+1} \rightarrow [0,1]$, 
returning the probability of transitioning from a state $\state_t$ to $\state_{t+1}$ given a joint-action profile~$\bm{\action}$;
a set of joint observations $\observations$; 
an observation probability function defined as
$\observationf_p : \states \times \actions_1 \times ... \times \actions_n \times \observations \rightarrow [0,1]$;
a discount rate $\gamma$,
and; for each player $p$, a reward function 
$\mathcal{\rewards}_p : \mathcal{\states}_t \times \mathcal{\actions}_1 \times ... \times \mathcal{\actions}_n \times \mathcal{\states}_{t+1} \rightarrow \realnumber$,
returning a reward~$r_p$. 
%
%
%
%

\subsection{The Adversarial Learning Challenge} \label{sec:adv_learning_challenge}

Our focus is on adversarial learning scenarios that feature
cyber-defence (\blue) and cyber-attacking (\red) agents.
Formally, for each agent (player) $p$, 
the policy $\pi_p$ is a mapping from the 
state space to a probability distribution over actions,
$\pi_p : \states_p \rightarrow \Delta(\actions_p)$.
Transitions within POMGs are determined by a joint policy $\bm{\pi}$.
%
%
Joint policies excluding agent~$p$ are defined as $\bm{\pi}_{-p}$. 
The notation $\langle \pi_p, \bm{\pi}_{-p}\rangle$ refers to a joint policy with 
agent $p$ following $\pi_p$ while the other agents follow $\bm{\pi}_{-p}$.
As noted above, the stationarity of an attacking agent's policy is a strong assumption. 
Therefore, selecting defending agents based on their performance against
a pool of known attackers runs the risk of being blindsided 
by an unfamiliar attacker. 
Here, a more desirable solution concept commonly used in this class of games is 
the Nash equilibrium~\citep{nash1951non}: 
\begin{definitionsec}[Nash Equilibrium]
A joint policy $\bm{\pi}^*$ is a Nash equilibrium \emph{iff} no player $p$ can improve their gain through unilaterally deviating from $\bm{\pi}^*$:
\begin{equation} \label{eq:nash_equilibrium}
\forall p, 
\forall \pi_p, 
\G{p}{\JointPolicy{\pi^*_p}{\bm{\pi}^*_{-p}}}
\geq  \G{p}{\JointPolicy{\pi_p}{\bm{\pi}^*_{-p}}}.
\end{equation}
\end{definitionsec}
Our focus is on finite two-player zero-sum games, 
where an equilibrium is referred to as a saddle point,
representing the \emph{value of the game}~$v^*$. 
Given two policies $\pi_1$, $\pi_2$, the equilibrium of a finite zero-sum game is:
\begin{theoremsec}[Minmax Theorem] \label{theorem:minmax} 
In a finite two-player zero-sum game
$v^* = max_{\pi_1} min_{\pi_2}\G{i}{\JointPolicy{\pi_1}{\pi_2}} = 
min_{\pi_2} max_{\pi_1} \G{i}{\JointPolicy{\pi_1}{\pi_2}}$.
\end{theoremsec}
The above theorem 
%
states that every finite, zero-sum, two-player game 
has optimal mixed strategies~\citep{v1928theorie}. 
%
%
%
However, in complex games, 
including ACD scenarios, 
finding a Nash equilibrium is typically intractable
due to the complexity of the strategy space for computing 
\emph{actual} best responses. 
Here the concept of an approximate Nash equilibrium ($\epsilon$-NE) is 
helpful~\citep{oliehoek2018beyond}: 
\begin{definitionsec}[$\epsilon$-Nash Equilibrium]
$\bm{\pi}^*$ is an $\epsilon$-NE \emph{iff}:
\begin{equation} \label{eq:eqp_ne}
\forall i, 
\forall \pi_i, 
\G{i}{\JointPolicy{\pi^*_i}{\bm{\pi}^*_{-i}}} \geq  
\G{i}{\JointPolicy{\pi_i}{\bm{\pi}^*_{-i}}} - \epsilon.
\end{equation}
\end{definitionsec}

Assuming an $\epsilon$-Nash Equilibrium has not yet been found,
our interest turns towards measuring the difference in the gain that 
an agent $j$ can achieve through unilaterally deviating from the 
current joint-policy profile.
This metric is known as \emph{exploitability}, and can be 
computed using the following equation~\citep{lanctot2017unified}:
\begin{equation} \label{eq:exploitablity}
\exploitabilityi = \G{j}{\JointPolicy{\pi_{i}}{\pi_{j}'}} -  \G{j}{\JointPolicy{\pi_{i}}{\pi_{j}}}.
\end{equation}
In the above equation $\exploitabilityi$ is the exploitability for agent~$i$, 
and~$\pi_{j}'$ is an approximate best response policy against~$\pi_{i}$.
If $\exploitabilityi \leq 0$, then $\pi_{j}'$ is unable to improve on~$\pi_{j}$. 
As a result~$\pi_{i}$ cannot be exploited.  
%
%
If the combined exploitability in a two-player zero-sum game is less than or equal to~$\epsilon$, 
$\Exploitability = \exploitabilityi + \exploitabilityj \leq \epsilon$, then players $i$ and $j$ are in an 
$\epsilon$-NE.

\subsection{Approximate Double Oracles} \label{sec:ado}

One of the long-term objectives of adversarial learning is to limit the 
exploitability of agents deployed in competitive environments~\citep{lanctot2017unified}.
While a number of theoretically grounded methods exist for limiting exploitability, 
computing the value of a game remains challenging in practise, 
even for relatively simple games. 
%

Here, we provide a recap of a popular,
principled adversarial learning frameworks
for finding a \emph{minimax} equilibrium
and reducing exploitability:
the DO algorithm~\citep{mcmahan2003planning}.
This algorithm defines a two-player zero-sum normal-form game~$\NormalFormGame$, 
where actions correspond to policies available to the 
players within an underlying 
(PO)MG~$\ExtensiveFormGame$.  
Payoff entries within~$\NormalFormGame$ are determined
through computing the gain~$\mathcal{G}$ for each policy pair
within~$\ExtensiveFormGame$:
\begin{equation}\label{eq:cell_entries}
\mathcal{R}^{\NormalFormGame}_{i}(\langle \action^r_1, \action^c_2\rangle) = 
\mathcal{G}^{\ExtensiveFormGame}_{i}(\JointPolicy{\pi^r_1}{\pi^c_2}).
\end{equation}
In \autoref{eq:cell_entries}, $r$ and $c$ refer to the respective
rows and columns inside the normal-form (bimatrix) game. 
%
The normal-form game~$\NormalFormGame$ is subjected 
to a game-theoretic analysis, to find an optimal mixture over actions
for each player,
%
representing a probability distribution 
over policies for the game~$\ExtensiveFormGame$. 

The DO algorithm assumes that both players
have access to a \emph{best response oracle} 
that returns a \emph{best response policy} $\pi_i$
for agent $i$ against 
a \mop played by the opponent: $\pi_i \gets O_i(\mu_j)$.\footnote{We discuss the benefits of mixed strategies in \autoref{appendix:benefits_of_mixtures}.}
Here, $\mu_j$ defines a sampling weighting over policies 
$\Pi_j$ available to agent $j$.  
%
%
%
%
Best responses are subsequently added to the list of available policies
for each agent.
As a result each player has an additional action 
that it can choose in the normal-form game~$\NormalFormGame$. 
Therefore, $\NormalFormGame$ needs to be augmented through 
computing payoffs for the new row and column entries.
Upon augmenting $\NormalFormGame$ another game-theoretic analysis is conducted, and the steps 
described above are repeated.
If no further best responses can be found, then the DO algorithm 
has converged upon a minimax equilibrium~\citep{mcmahan2003planning}.

Learning an \emph{exact} best response is often intractable.
In games that suffer from the curse-of-dimensionality,
$O(\mu)$ will often only return an 
\emph{approximate best response}, 
giving rise to the \emph{approximate} DO algorithm 
%
%
(ADO, Algorithm~\ref{alg:ADO}).
This approach is guaranteed to converge upon an \emph{$\epsilon$-resource bounded NE} ($\epsilon$-RBNE)~\citep{oliehoek2018beyond}.
\begin{algorithm}[H]
\caption{The Approximate Double Oracle Algorithm}
\label{alg:ADO}
\providecommand{\commentSymb}{//}
\begin{algorithmic}[1]
\small
\STATE{$\langle \aA{\blue}, \aA{\red} \rangle \gets \funcName{InitialPolicies}()$}
\STATE{$\langle \mA{\blue}, \mA{\red} \rangle \gets \langle \{\aA{\blue}\}, \{\aA{\red}\} \rangle$} \COMMENT{Set initial mixtures}
\WHILE{True}
    \STATE{$\aA{\blue} \gets O_{\blue}( \mA{\red})$}  
    \STATE{$\aA{\red} \gets O_{\red}( \mA{\blue} )$}
    \STATE{$\Exploitability \gets \G{\blue}{\aA{\blue},  \mA{\red} } + \G{\red}{\mA{\blue}, \aA{\red}} $}  
    \IF{$\Exploitability \leq \epsilon$}
	\STATE \textbf{break} \COMMENT{Found $\epsilon$-RBNE}
    \ENDIF
    \STATE{$\mathcal{N} \gets \funcName{AugmentGame}(\mathcal{N}, \aA{\blue}, \aA{\red})$}
    \STATE{$\langle \mA{\blue}, \mA{\red} \rangle \gets \funcName{SolveGame}(\mathcal{N})$}
\ENDWHILE
\end{algorithmic}
\end{algorithm}
\vspace*{-2em}
%
%
%
%
%
%
%
\subsection{Potential-Based Reward Shaping}

The iterative learning of ABRs presents a computationally costly endeavour,
highlighting the need for methods that expedite the search for ABRs~\citep{liuneupl}.
We observe that learning ACO agents are confronted
with the \emph{temporal credit assignment problem}. 
For example, in numerous cyber-defence scenarios, large penalties
are associated with ACA agents impacting high-value targets, 
such as operational hosts and servers~\citep{cyborg_acd_2021}.
However, attacks on an operational subnet will typically be 
preceded by Blue/Red actions to protect/compromise more accessible
user and enterprise subnets.      
Here, reward shaping offers a means of
mitigating the temporal credit assignment 
problem via a modified reward function $\rewards' = \rewards + F$, 
where $F$ represents the shaping reward~\citep{grzes2009theoretical}.

A shaping reward can be implemented using 
domain knowledge or learning approaches~\citep{grzes2009learning}.
However, unprincipled reward shaping can lead to policies that behave
(near)-optimally under $\rewards'$ while performing sub-optimally with respect to $\rewards$~\citep{ng1999policy}.
Classic examples include agents learning to circle around a goal location
to maximise a distance-based shaping reward~\citep{randlov1998learning} 
and soccer agents losing interest in scoring goals upon receiving a shaping
reward for successfully completing passes~\citep{ng1999policy}. 

To address the above challenge, \cite{ng1999policy} proved 
that $F$ being a \emph{potential-based shaping function}, 
\begin{equation} \label{eq:pbrs}
F(s, s') = \tau(\gamma \Phi(s') - \Phi(s)),
\end{equation}
is a necessary and sufficient condition to guarantee consistency when learning
an optimal policy on an MDP  
$\M' = (\states, \actions, \transition, \rewards + F, \gamma)$
instead of 
$\M = (\states, \actions, \transition, \rewards, \gamma)$. 
%
Here, 
$\Phi : \states \rightarrow \realnumber$ is a real-valued \emph{potential function}, 
defined over source and destination states, 
and $\tau > 0$ is used to scale the shaping reward~\citep{grzes2009theoretical}.
Crucially, PBRS's convergence 
guarantees hold even when using a suboptimal potential function~\citep{gao2015potential}.

%
%
%
%
%

\section{Methods} \label{sec:methods}

\subsection{Value-Function Potential-Based Reward Shaping} \label{sec:methods:vfpbrs} 

%
%
%
%

In their seminal work on policy invariance under PBRS transformations,
\cite{ng1999policy} state that one way for defining a good
potential function $\Phi$ is to approximate the optimal value function (VF)
$V^*_\M(s)$ for a given problem~$\M$.
We observe that any \ADO run featuring oracles using
VFs can provide a plethora of approximate VFs, $V_{\M_k}(s)$,
enabling a VF driven PBRS (VF-PBRS);
with $\M_k$ representing the (PO)MG that an ABR $k$ was trained on.

The number of VFs available 
after multiple ABR iterations raises the 
question of which $V_{\M_k}$ to select. 
Here, the ADO algorithm's
current mixture~$\mu_i$ captures the best
response for player~$i$ using available policies 
within our empirical matrix game against
player~$j$'s mixture~$\mu_j$. 
We hypothesise that~$\mu_i$ can 
also help us select informative VFs.  

Rather than sampling a single~$V_{\M_k}$
we propose using a weighted ensemble.
However, normalization is required when ensembling VFs,
as the magnitudes of approximated value estimates are 
often not directly comparable~\citep{garcia2024online}.
We address this by applying Z-score normalization to each VF: $Z(V_{\M_k}(s))$.
%
The resulting potential function is the weighted sum of normalised value estimates:
\begin{equation} \label{eq:pbrs:ensemble}
\Phi(s) =  \sum_{k=1}^{\left|\mu\right|} \mu_i^k \times Z(V_{\M_k}(s)).
\end{equation}

\textbf{Theoretical Analysis:} Oracles using VF-PBRS do not impact the
theoretical guarantees underpinning the DO algorithm.
Given a two-player zero-sum game, 
a Nash equilibrium has been found 
when neither player can compute a best response
that outperforms their respective current mixture:
\begin{equation}
\begin{aligned} 
\Exploitability & = 
\G{i}{\JointPolicy{\O{i}{\mu_j}}{\mu_j}} +  
\G{j}{\JointPolicy{\mu_i}{\O{j}{\mu_i}}} \\
 & \leq 0 = \G{i}{\JointPolicy{\mu_i}{\mu_j}} + 
\G{j}{\JointPolicy{\mu_i}{\mu_j}}.
\label{eq:stopcondition}
\end{aligned}
\end{equation}
It follows that $\Exploitability \leq 0$ relies on $\O{i}{\mu_j}$ 
and $\O{j}{\mu_i}$ returning \emph{best responses}. 
Below we show that oracles using VF-PBRS meet this requirement in Markov games.
\begin{theoremsec} \label{def:vfpbrs_with_do}
Given a best response oracle $O_{i}$ that uses a 
shaping reward $\rewards' = \rewards + F$, 
then $F$ being a potential-based reward shaping function is a \emph{sufficient} 
condition to guarantee that $\pi_i \gets O_{i}(\mu_j)$  is a 
best response for $\M' = (n, \states, \actions, \transition, \rewards + F, \gamma)$
and $\M = (n, \states, \actions, \transition, \rewards, \gamma)$. 
\end{theoremsec}
\begin{proof}
%
%
The DO algorithm requires that 
the opponent policies sampled from the mixture $\mu_j$ remain stationary while 
player $i$ computes a best response.
Given that player $j$ is treated as part of
the environment under this formulation, both $\M$ and $\M'$ are equivalent to
Markov decision processes (MDPs) $\M = (\states, \actions, \transition, \rewards, \gamma)$
and $\M' = (\states, \actions, \transition, \rewards + F, \gamma)$.
\cite{ng1999policy}'s proof of sufficiency\footnote{Included in Appendix \ref{app:proof_of_sufficiency} for convenience.} 
shows that if $F$ is a potential-based shaping function,~then:
\begin{equation}
\pi^*_{\M'}(s) = \argmax_{a \in \actions} Q^*_{\M'}(s,a) = \argmax_{a \in \actions} Q^*_{\M}(s,a),
\end{equation}
with $Q^*$ representing the optimal Q-value for state $s$ and action~$a$.
Therefore, $\pi^*_{\M'}$ represents a best response against the opponent mixture $\mu_j$
in both $\M'$ and $\M$.
\end{proof}

The complexity of our target domain requires oracles that rely on function approximators.
Therefore, the best we can hope for are ABRs.
Here, \cite{ng1999policy} observe that PBRS is robust in the sense 
that near-optimal policies are also preserved.
Given a PBRS function $F$, any near-optimal policy 
learnt in $\M'$ will also be a near-optimal policy in $\M$, 
meaning that an ABR policy $\pi$ obtained using VF-PBRS,
with $\left|V^\pi_{\M'}(s) - V^{\pi*}_{\M'}(s)\right| < \varepsilon$,
will also be near optimal in $\M$: $\left|V^\pi_{\M}(s) - V^{\pi*}_{\M}(s)\right| < \varepsilon$.
Therefore, \ADO's theoretical guarantees for converging upon an 
$\epsilon$-RBNE remain intact.  

\subsection{Multiple Response Oracles} \label{sec:multi_oracle} 

One of the disadvantages of the \ADO 
formulation is that the 
function $O_i(\mu_j)$ only returns a single policy~$\pi_i$ 
per ABR iteration.
What if we have multiple approaches for computing
responses against an opponent mixture, and do not know
in advance which one is most likely to return the \emph{ABR}?
%
%
Here, we propose a novel \ADO formulation to address this limitation. 
%
We call this approach the \emph{multiple response oracles} algorithm (\MRO).
In contrast to the \ADO algorithm, \mro replaces ABR 
oracles $O_i$ with functions $\Pi_i \gets R_i(\mu_j)$,
that return a set of responses.
We note that this extension does not change the underlying theoretical 
properties of the \ADO algorithm. 
In the set of response policies
computed by agent~$i$ against the opponent mixture $\mu_j$, 
using a responses function $R_i(\mu_j)$, 
there will exist an ABR policy $\pi^{*}_{i}$, where: 
\begin{equation} \label{eq:mro:abr_policies}
\G{i}{\pi^{*}_{i},  \mA{j}} \geq \G{i}{\pi_{i},  \mA{j}}, \forall \pi_{i} \in R_i(\mu_j). 
\end{equation}
Therefore, assuming an oracle $O_i$, that can identify 
the ABR policy as per \autoref{eq:mro:abr_policies}, we have:
%
$\pi^{*}_{i} \gets O_i(R_i(\mu_j))$ (See Appendix~\ref{sec:multi_oracle:theory}
for a detailed theoretical analysis). 
\begin{algorithm}[h]
\caption{Multiple Response Oracles}
\label{alg:MRO}
\providecommand{\commentSymb}{//}
\begin{algorithmic}[1]
\small
\STATE{$\langle \aA{\blue}, \aA{\red} \rangle \gets \funcName{InitialPolicies}()$}
\STATE{$\langle \mA{\blue}, \mA{\red} \rangle \gets \langle \{\aA{\blue}\}, \{\aA{\red}\} \rangle$} \COMMENT{set initial mixtures}
\WHILE{True}
    \STATE{$\langle \pi_{\blue}^0, ..., \pi_{\blue}^n \rangle \gets R_{Blue}( \mA{\red})$}  
    \STATE{$\langle \pi_{\red}^0, ..., \pi_{\red}^m  \rangle \gets R_{Red}( \mA{\blue} )$}
    \STATE{$\Exploitability \gets \G{\blue}{\pi^{*}_{\blue},  \mA{\red} } + \G{\red}{\mA{\blue}, \pi^{*}_{\red}} $} 
    \IF{$\Exploitability \leq \epsilon$}
	\STATE \textbf{break} \COMMENT{found $\epsilon$-RBNE}
    \ENDIF
    \STATE{$\mathcal{N} \gets \funcName{AugmentGame}(\mathcal{N}, \pi_{\blue}^{0, ...,  n}, \pi_{\red}^{0, ..., m})$}
    \STATE{$\langle \mA{\blue}, \mA{\red} \rangle \gets \funcName{SolveGame}(\mathcal{N})$}
\ENDWHILE
\end{algorithmic}
\end{algorithm}
\vspace*{-1em}
%
\subsection{Pre-trained Model Sampling} \label{sec:methods:ptm_sampling}

Training ACO agents 
requires lengthy wall-times to ensure an exhaustive exploration 
of semantically similar state-action pairs.
To address this challenge, a general 
consensus has emerged within the AI community
that PTMs should be utilised when
possible, rather than training models from scratch~\citep{han2021pre}. 
Within the context of DO-based approaches, 
agents have the luxury of an iteratively expanding pool from which PTMs
can be sampled. 
Here, once again, the mixtures~$\mu_i$ can provide guidance;
on this occasion for sampling from PTMs that represent the best \emph{available} response against 
the current opponent mixture~$\mu_j$.
However, one of the strengths of the DO algorithm is the exploration 
of the strategy space~\citep{wellman2024navigating}. 

We propose balancing the above trade-off by adding $\epsilon$-greedy exploration
to our oracles. 
In each ABR iteration a mixture-guided PTM is sampled 
with a probability $1 - \epsilon$.
Initially $\epsilon$ is set to $1$. 
A decay rate $d \in [0, 1)$ is subsequently applied after each ABR iteration.
Exploratory iterations can either consist of training 
a freshly initialised agent, or using a designated PTM as a starting
point, \eg a policy that can be classed as a \emph{generalist}. 
In contrast, in greedy iterations the mixture weights are used as sampling probabilities.
%

\section{Evaluation Environments} \label{sec:envs}

Below we conduct an empirical game-theoretic evaluation of
DRL approaches in two post-exploitation lateral movement scenarios:
CybORG CAGE Challenges 2 and~4~\citep{cage_challenge_2_announcement,cage_challenge_4_announcement}. 
Network diagrams and experiment settings can be found in \autoref{app:network_diagram} and \ref{app:experiment_settings}.

\textbf{CAGE Challenge 2 (CC2):} A cyber-defence agent (Blue) is tasked
with defending a computer network containing key manufacturing 
and logistics servers distributed across three subnets: \texttt{user}, \texttt{enterprise} and \texttt{operations}.
Each episode begins with an attacking agent (Red) having root access on a 
user host on the \texttt{user} subnet. 
Red's objectives are to reach an operational server and degrade network services. 
Blue’s objective is to minimise the presence of the Red agent and ensure that network
functionality is maintained through running detailed analyses on hosts, launching
decoy services, and removing malicious software~\citep{kiely2023autonomous}. 
For cases where the Red agent is too well established,
Blue can also resort to restoring hosts using a clean backup. 

\textbf{CAGE Challenge 4 (CC4):} 
Blue is tasked with defending 
sub-networks
responsible for supporting military operations. 
Red's goal is to disrupt
a base-station from which Blue 
 unmanned aerial vehicle activities are coordinated. 
The network in CC4 is highly segmented to increase the level of security.
Therefore, a multi-agent (MA) cyber-defence solution is required. 
Here each agent is assigned
a zone that it must protect within two high value \texttt{Deployed Networks} (A and B).
Each network has a \texttt{Restricted} and \texttt{Operational} zone.
Successful Red attacks on these zones result in large penalties for Blue. 

\section{Cyber-Defence \& Attacking Oracles} \label{sec:envs}
\textbf{Cardiff University (Blue):}
Our first Blue oracle is a modified version of the CC2 winning submission
from Cardiff University~\citep{JohnHannay}.
This approach, based on the Proximal Policy Optimization (PPO) algorithm~\citep{schulman2017proximal}, 
benefits from a significantly reduced action space.
It defines a single decoy action per host 
that greedily selects from available decoys.
The submission also uses 
a \say{fingerprinting}
function for identifying the two rules-based Red agents supplied by CC2
and countering with respective ABR policies.
%
%
However, fingerprinting new Red policies is beyond our current scope.
We therefore disable this method. 
%
%
%
%

\textbf{Cybermonic Inductive Graph-PPO (GPPO, Blue):}
Given that cyber-defence scenarios typically feature underlying dynamics
unique to graph-based domains, ACD agents are increasingly benefitting from  
incorporating graph machine learning approaches~\citep{10.1145/3689933.3690835}. 
\cite{CC2KEEP} and \cite{CC4KEEP} have implemented two OS Graph-PPO algorithms
for CC2 and CC4.
The approaches utilise graph observation wrappers for converting
CybORG's observations into an augmented graph environment. 
An internal state is maintained that keeps track of changes in the graph structure.
Graph convolutional networks~\citep{kipf2016semi} are used to extract features from the graph representation.
%
%
%
An \emph{independent learner} (IL) approach is used in CC4, 
with each agent updating independent actor-critic networks using local observations. 

\textbf{Action Masking PPO (AM-PPO, Red):}
The cyber-attacking agents within CC2\&4 must also
cope with a high-dimensional action space.
Therefore, to confront the Blue agents with challenging opponents, 
our cyber-attacking agent has the benefit 
of being able to use action masking during training.
This was enabled by adding a Red action-masking wrapper
to CC2, and implementing (MA)DRL action masking agents. 
For CC4 our AM-MAPPO agents also use an IL approach. 
We set the Red rewards to the negation of the Blue rewards in both environments.

\section{Empirical Game-Theoretic Analysis} \label{sec:results}

%

\subsection{CybORG CAGE Challenge 2} \label{sec:results:cc2}

During lengthy preliminary runs using 5M time-steps per response, 
it was observed that the agents would often make significant improvements after 
2.5M time-steps (See \autoref{appendix:additional_evals:cc2}).
However, this corresponds to approximately 
one day per ABR iteration.
To reduce wall-time, we instead
initialise from \say{generalist} PTMs
obtained from previous 5M step runs. 
We allow 1.5M environment steps per response. 
Responses are evaluated using the mean episodic reward
from 100 evaluation episodes. 

\begin{figure}[H]
\includegraphics[trim={0cm 0.5cm 0cm 0.5cm},width=1.0\columnwidth]{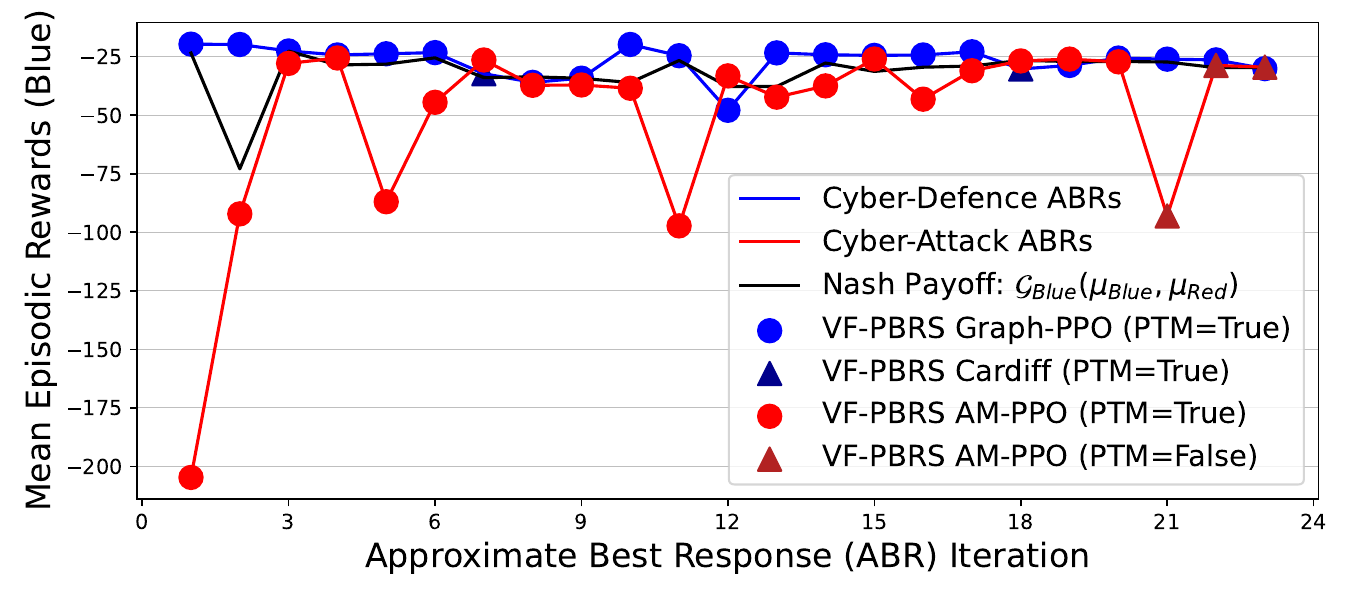}
\caption{A depiction of Blue and Red ABRs from an MRO run on CC2.
Rewards are plotted from Blue's perspective.
%
%
Both agents are unable to find ABRs that significantly improve 
on the Nash payoff (the black line) between iterations 11 -- 20.
Red oracles use PTMs for the first 20 iterations, 
but subsequently switch to random initialisations (PTM=False). 
Initially this oracle setting finds a policy that 
significantly improves on the Nash payoff.
However, the Blue agent immediately counters 
through adjusting its mixture.
}
\label{fig:cc2_convergence}
\end{figure}

\noindent{\textbf{Impact of using PTMs and VF-PBRS:}} 
Upon completing around 20 ABR iterations, we find that Blue and Red 
oracles using PTMs struggle to improve on 
the Nash payoffs under the joint mixture profile 
$\langle \mu_{Blue}, \mu_{Red} \rangle$
(see \autoref{fig:cc2_convergence}).
At this point we confront the Blue mixture  
with a Red oracle that uses a random initialisation
and a budget of 5M time-steps.
This Red oracle initially finds an ABR, $\pi_{Red}^{21}$, that 
significantly improves upon
the Nash payoff, from $27.30$ under $\langle \mu_{Blue}, \mu_{Red}\rangle$ 
to $93.05$ for $\langle \mu_{Blue}, \pi_{Red}^{21} \rangle$. 
However, upon adding $\pi_{Red}^{21}$ to the empirical game,
the Blue agent neutralises the new Red TTPs through adjusting
its mixture, decreasing the Nash payoff to $29.75$.
The Red agent is subsequently unable to find a 
significantly better response.

We continue to probe the final mixtures, 
seeking oracle settings that outperform
the responses from the final ABR iteration.
However, the resulting responses do not significantly
deviate from the final value of the game estimate, $\G{Blue}{\mu_{Blue}, \mu_{Red}}=-29.63$. 
Nevertheless, our evaluation yields interesting insights into the benefits of VF-PBRS.
For instance, when comparing \say{full} (PTM=False, 5M time-step) 
training runs, VF-PBRS responses outperform the corresponding vanilla 
responses (see \autoref{fig:vfpbrs_table}). 

Gathering a sufficient number of 5M step runs to test the statistical 
significance of the above result is a computationally expensive endeavour.
Instead we initialise from PTMs and gather ten 1.5M step runs per setting. 
%
%
We compare two VF-PBRS configurations ($\tau = \{1, 0.5\}$) against vanilla responses.
For Red we find that 
on average VF-PBRS ($\tau = 1$: $24.72$, and; $\tau = 0.5$: $24.85$)
marginally outperform the vanilla responses ($24.22$).
However, the differences are not significant.
This result also shows that the generalist policy 
$\pi^{G}_{Red}$ is a poor PTM choice against advanced Blue mixtures,
when compared against the results from the full runs.

%
In contrast, upon conducting the same
experiment with Blue GPPO responses,
both VF-PBRS using VF ensembling (VFE) and
individual VFs significantly outperform the vanilla 
responses (see \autoref{fig:vfpbrs_scatter_blue}).
Using VFE outperforms the majority of runs conducted 
with individual VFs (all using $\tau=1$).
Finally, no significant difference is found when using a VFE setting of $\tau=0.5$,
compared to the VFE runs with $\tau=1$.
%
%
\begin{table}[H]
\footnotesize
\resizebox{\columnwidth}{!}{
\begin{tabular}{|c|c|c|c|c|}
\hline 
\textbf{Agent} & \textbf{PTM} & \multicolumn{2}{|c|}{\textbf{Description}} & \textbf{Payoffs} \\
\hline 
\hline 
\multirow{12}{2em}{\textbf{\textcolor{blue}{Blue}}} 
& \multirow{8}{2em}{\textbf{True}} 
& \multirow{5}{*}{VF$k$ ($\mu_i^k$)} 
&    VF1 ($0.1$)    & $-33.8$ ($\pm 3.7$) \\
\cline{4-5}
& & & VF2 ($0.2$) & $-32.1$ ($\pm 1.0$) \\ 
\cline{4-5}
& & & VF3 ($0.3$)& $-34.0$ ($\pm 3.9$) \\ 
\cline{4-5}
& & & VF4 ($0.1$) & $-31.0$ ($\pm 1.7$) \\ 
\cline{4-5}
& & & VF5 ($0.3$) & $\bm{-29.4}$ \textbf{($\bm{\pm 0.5}$)} \\
\cline{3-5}
& & \multirow{2}{*}{VFE ($\tau$)} 
& $\tau=1$ & $-29.8$ ($\pm 1.2$) \\ 
\cline{4-5}
& & & $\tau=0.5$ & $-31.3$ ($\pm 1.2$) \\
\cline{3-5}
& & Vanilla & $\tau=0$ & $-38.3$ ($\pm 2.7$) \\
\cline{2-5}
& \multirow{4}{2em}{\textbf{False}} 
& \multirow{2}{*}{VFE ($\tau=1$)} 
& GPPO & $\bm{-27.8}$ ($\pm 1.5$) \\
\cline{4-5}
& & & CAR & $-30.3$ ($\pm 1.5$) \\
\cline{3-5}
& & \multirow{2}{*}{Vanilla} 
& GPPO & $-29.5$ ($\pm 1.7$)\\
\cline{4-5}
& & & CAR & $-31.2$ ($\pm 2.3$) \\
\hline
\hline
\multirow{5}{2em}{\textbf{\textcolor{red}{Red}}} 
& \multirow{3}{2em}{\textbf{True}} 
& \multirow{2}{*}{VFE ($\tau$)} 
& $\tau=1$ & $24.7$ ($\pm 0.4$) \\
\cline{4-5}
& & & $\tau=0.5$ & $\bm{24.9}$ \textbf{($\bm{\pm 0.5}$)} \\
\cline{3-5}
& & Vanilla & $\tau=0$ & $24.2$ ($\pm 0.5$) \\
\cline{2-5}
&  \multirow{2}{2em}{\textbf{False}} 
& VFE & $\tau=1$ & $\bm{29.5}$ ($\pm 0.8$)\\
\cline{3-5}
& & Vanilla & $\tau=0$ & $28.6$ ($\pm 0.4$) \\
\hline 
\end{tabular}}
\caption{A comparison of additional responses
learnt against the penultimate 
CC2 mixtures, following 
iteration~22.  
We compare the impact of using PTMs and VF-PBRS vs Vanilla training,
i.e., without reward shaping.
For VF-PBRS, we make the distinction between using VF ensembling (VFE) and
using individual VFs~$k$ with a mixture weighting $\mu_i^k > 0$.
Responses learnt without PTMs represent a single run per setting, while
ten runs were obtained for configurations using PTMs.
Listed are the mean and standard error upon conducting 100 evaluation episodes per response.}
\label{fig:vfpbrs_table}
\end{table}

\begin{figure}[H]
\includegraphics[width=1.0\columnwidth]{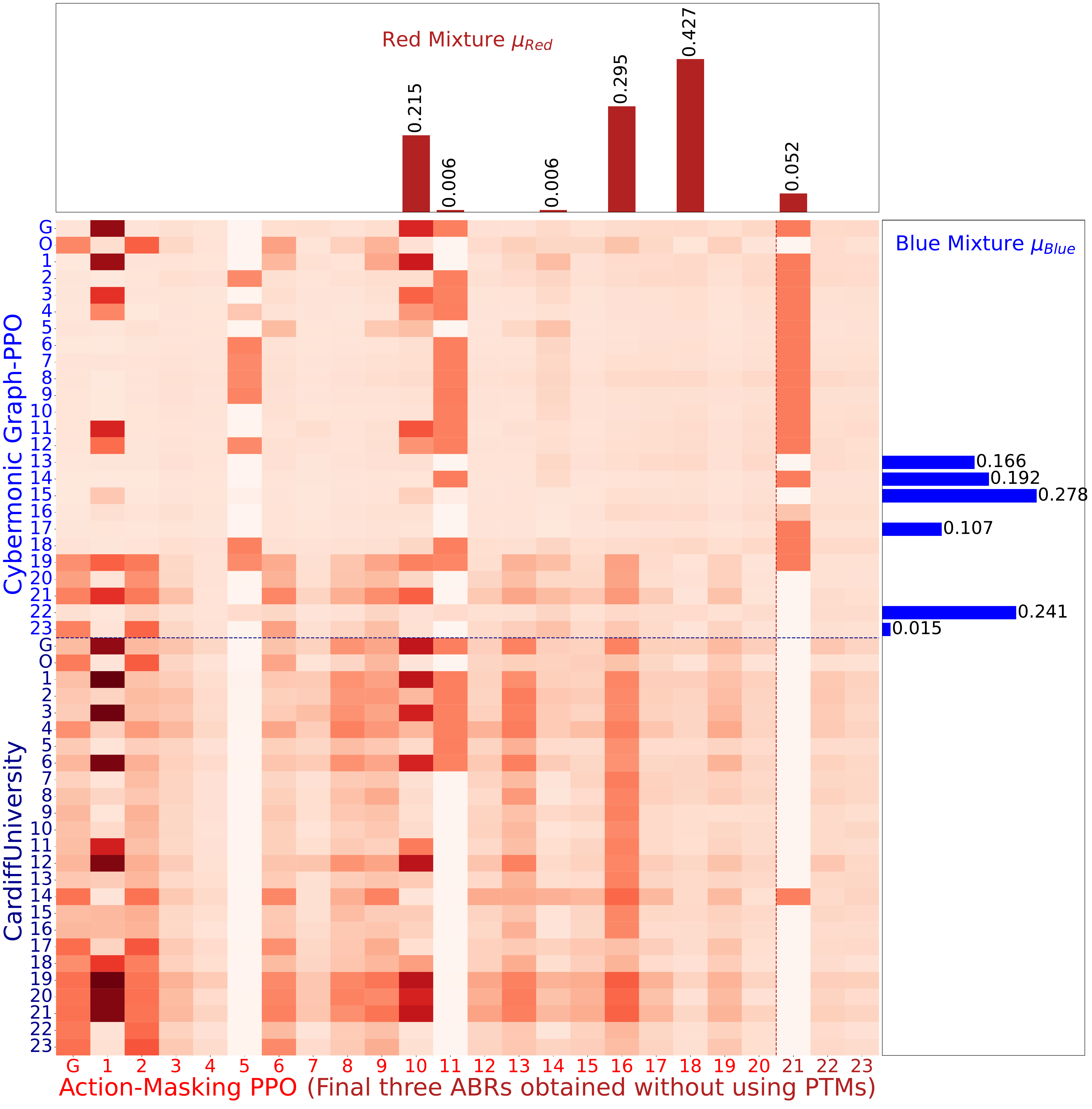}
\includegraphics[trim={0cm 0.5cm 0cm 0cm},width=1.0\columnwidth]{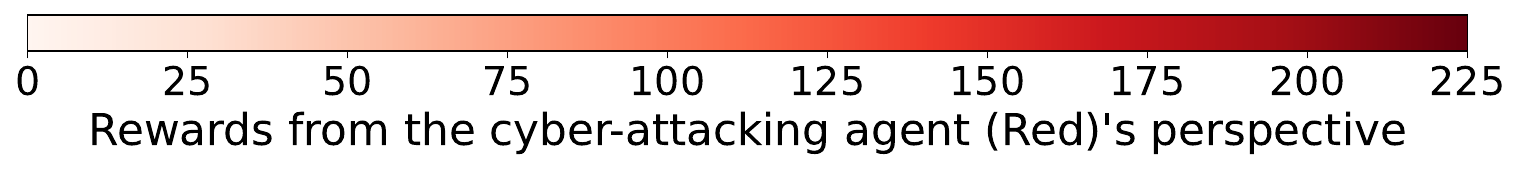}
\caption{An illustration of CC2's empirical game and Nash mixtures.
Cells represent the mean episodic reward for each policy pairing plotted from Red's perspective (100 evaluation episodes).
X and Y ticks indicate the ABR iteration in which a response was learnt.
We also include original (O) and generalist (G) policies. 
Darker cells represent match-ups that are favorable for Red.
%
%
}
\label{fig:cc2_payoff_matrix}
\end{figure}

{\parskip 0em
\begin{figure}[H]
\centering
\includegraphics[trim={0cm 1cm 0cm 1cm},width=\columnwidth]{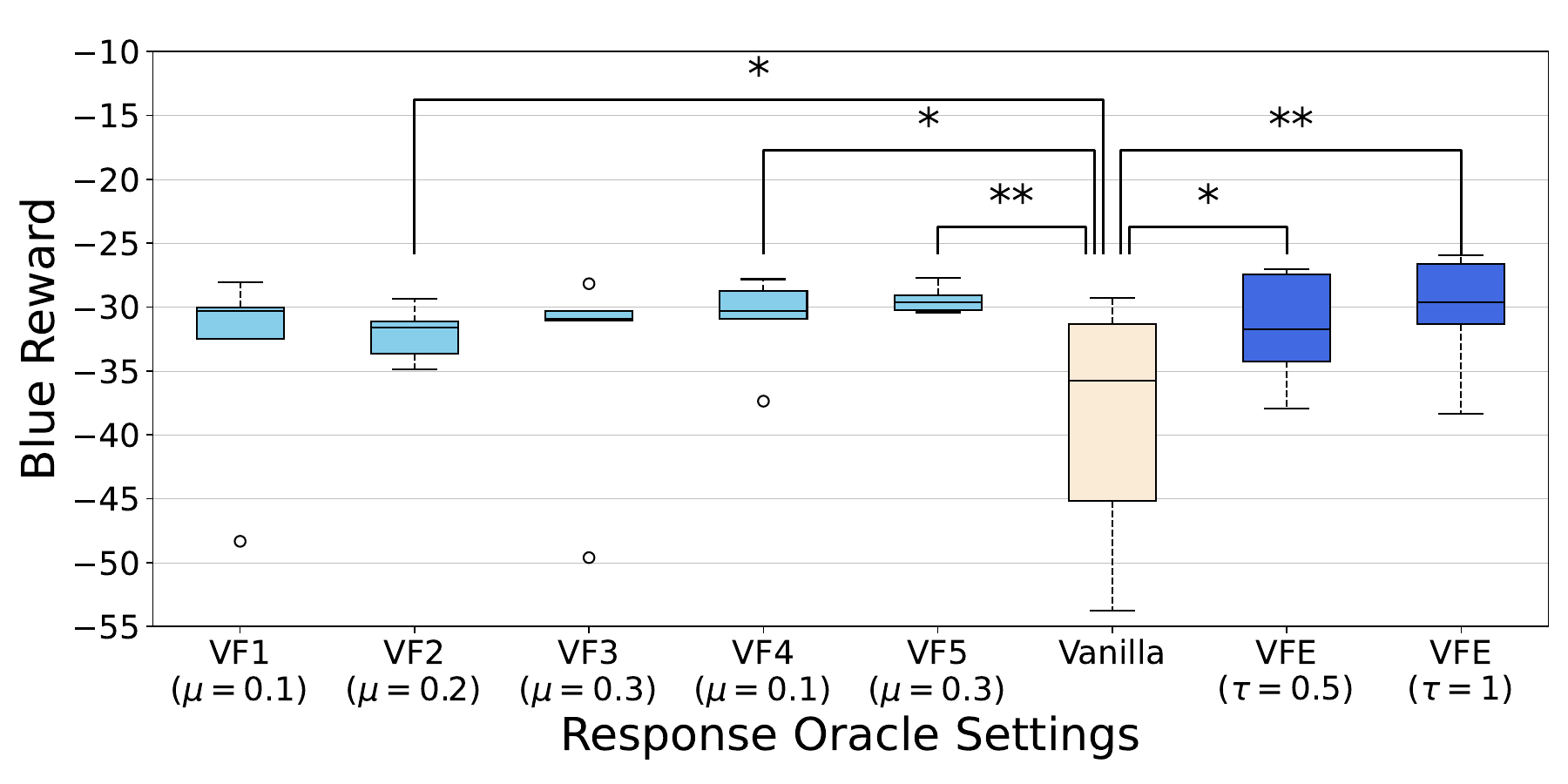}
\caption{The box plot above compares VF-PBRS runs using VF ensembling (VFE) and individual VFs~$k$ (where $\mu_i^k > 0$) 
against vanilla training runs (10 runs per setting). 
Statistical $p$-values of less than $0.05$ and $0.01$ are flagged with one and two asterisks respectively.
Outliers are plotted as separate black circles.
}
\label{fig:vfpbrs_scatter_blue}
\end{figure}
\noindent{\textbf{Final Mixture Composition:}}
\autoref{fig:cc2_payoff_matrix} illustrates the final empirical game 
for CC2.}
We observe that the final mixture composition for Blue consists of GPPO policies,
which often generalise well across learnt Red policies.
This naturally raises the question of how these policies differ from those not included.
Here, we compare the original Blue GPPO agent $\pi_{Blue}^o$, provided by \cite{CC2KEEP}, 
against the final Blue mixture agent $\mu_{Blue}$.
We evaluate $\pi_{Blue}^o$ against Red's best response,
$\pi_{Red}^2$: $\G{Blue}{\pi_{Blue}^o, \pi_{Red}^2} = -117.82$.
This attacking policy thrives against $\pi_{Blue}^o$
through gaining privileged access on \texttt{User2} and revealing
\texttt{Enterprise1}'s IP address~\footnote{CC2's network diagram is provided in \autoref{app:network_diagram}.}.
It subsequently launches  
\texttt{BlueKeep} and \texttt{EternalBlue} attacks against
\texttt{Enterprise1}, gaining privileged access on
$59.68\%$ of time-steps\footnote{These exploits give direct SYSTEM access to windows nodes. See the action tutorial in \cite{cage_challenge_2_announcement}.}.
This forces $\pi_{Blue}^o$ into repeatedly selecting the costly restore action.
In contrast, under $\langle \mu_{Blue}, \pi_{Red}^2 \rangle$ 
($\mathcal{G}_{Blue}=-21.96$)
the attacker is mostly restricted to the user 
network (See \autoref{fig:attacks_on_nodes}).\footnote{Additional policy characteristics are 
discussed in \autoref{appendix:additional results}.}
{\parskip 0em
\begin{figure}[H]
\centering
\includegraphics[clip,trim={6cm 0.9cm 6cm 0.9cm},width=\columnwidth]{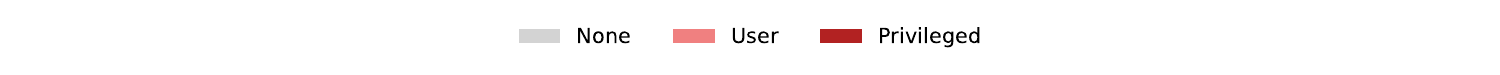}
\resizebox{\columnwidth}{!}{
\subfloat[$\langle \pi_{Blue}^o, \pi_{Red}^2 \rangle$]{
\label{fig:attacks_against_mixture_blue}
\includegraphics[clip,trim={0cm 0.8cm 0cm 0.2cm},width=0.5\columnwidth]{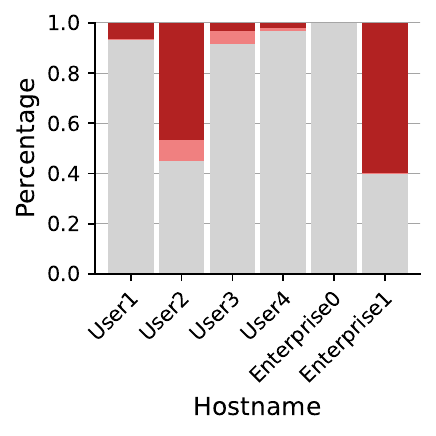}}
\hspace{0.3cm}
\subfloat[$\langle \mu_{Blue}, \pi_{Red}^2 \rangle$]{
\label{fig:attack_against_final_blue}
\includegraphics[clip,trim={0.8cm 0.8cm 0cm 0.2cm},width=0.45\columnwidth]{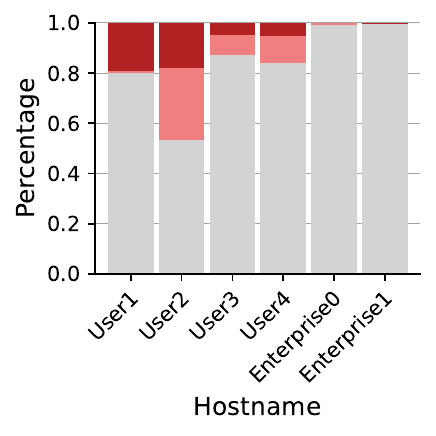}}
}
\caption{Depicted are the percentage of steps per episode where
Red can obtain \texttt{User} and \texttt{Privileged} access
on the listed nodes. 
We compare privileges obtained by Red against the original
GPPO parameterisation, $\pi_{Blue}^o$, and the
final mixture agent $\mu_{Blue}$.
The attacking policy is Red's ABR against $\pi_{Blue}^o$:  $\pi_{Red}^2$.}
\label{fig:attacks_on_nodes}
\end{figure}
\subsection{CybORG CAGE Challenge 4} \label{sec:cc4:results}}

For CC4 we launch an MRO run consisting of
2.5M step ABR iterations using PTMs.
Blue and Red each use two oracles: i.) VF-PBRS, and; ii.) Vanilla.
%
%
The MA-GPPO parameterisation $\bm{\pi}_{Blue}^o$ from~\cite{CC4KEEP}
represents the initial policy for Blue. 
For Red we compute an initial ABR against $\bm{\pi}_{Blue}^o$ 
for 5M time-steps, which serves as $\bm{\pi}_{Red}^o$. 
As with our CC2 experiment, we switch to \say{full} ABRs
once the MRO run with PTMs converges. 

%
%
%

%

%
%
%
%
%
\begin{figure*}
\centering
\begin{minipage}{\columnwidth}
\centering
\textbf{\footnotesize Restricted Zone A}
\end{minipage}
\begin{minipage}{\columnwidth}
\centering
\textbf{\footnotesize Operational Zone A}
\end{minipage}

\resizebox{\textwidth}{!}{
\subfloat[$\bm{\pi}_{Blue}^o$]{
\label{fig:cc4:action_percentages:rz:o}
	\includegraphics[trim={0cm 0 0cm 0},width=0.5\columnwidth]{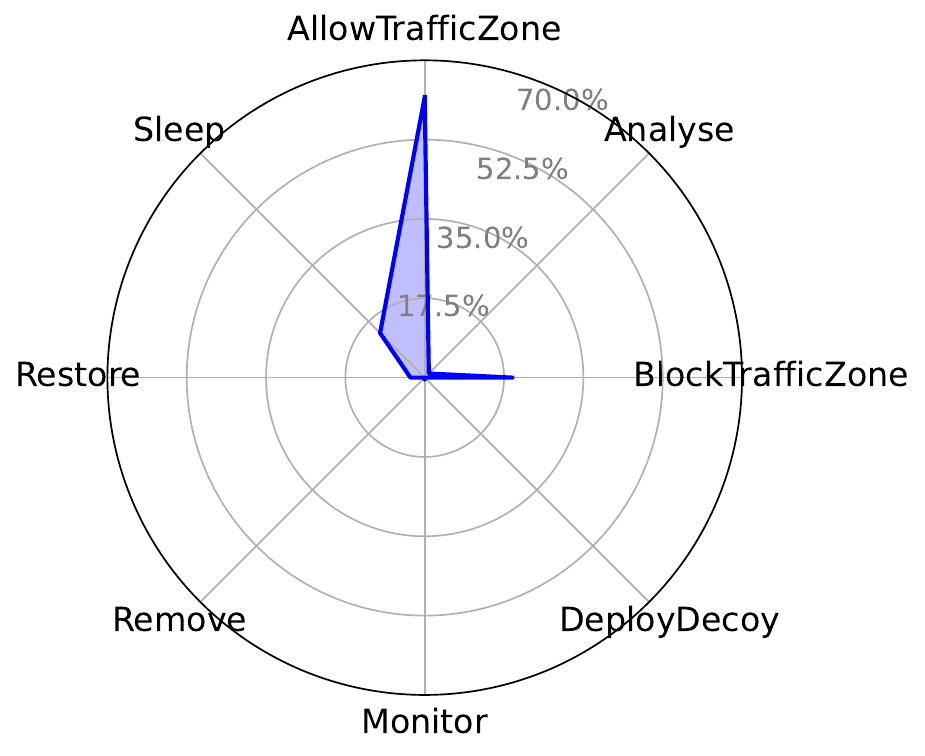}
}
\subfloat[$\bm{\pi}_{Blue}^1$]{
\label{fig:cc4:action_percentages:rz:abr1}
	\includegraphics[trim={0cm 0 0cm 0},width=0.5\columnwidth]{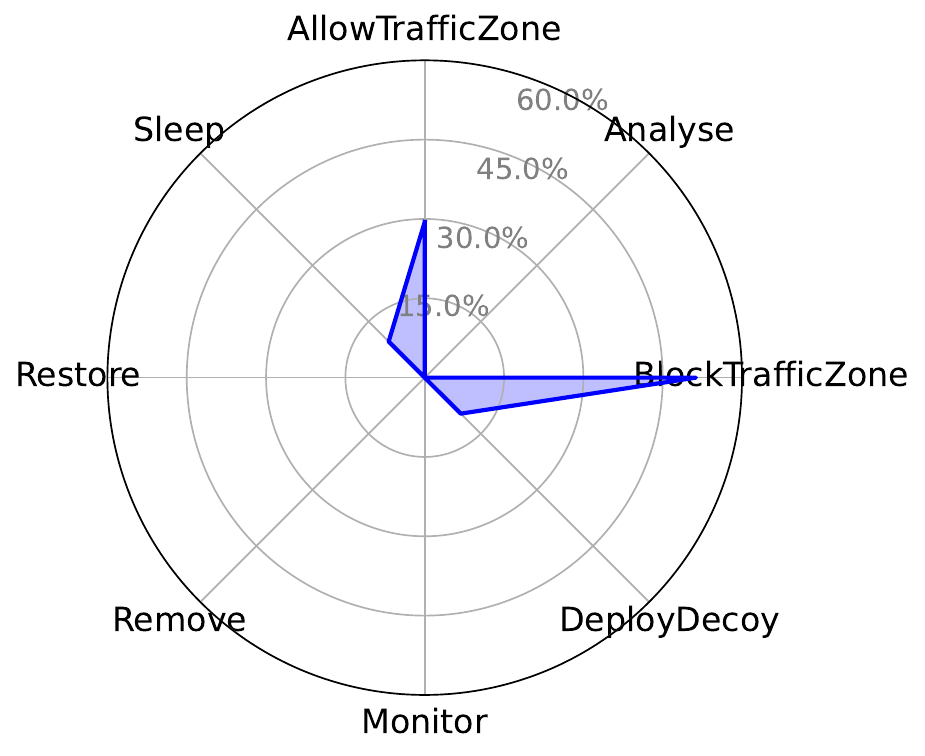}
}
\subfloat[$\bm{\pi}_{Blue}^o$]{
\label{fig:cc4:action_percentages:oz:o}
	\includegraphics[trim={0cm 0 0cm 0},width=0.5\columnwidth]{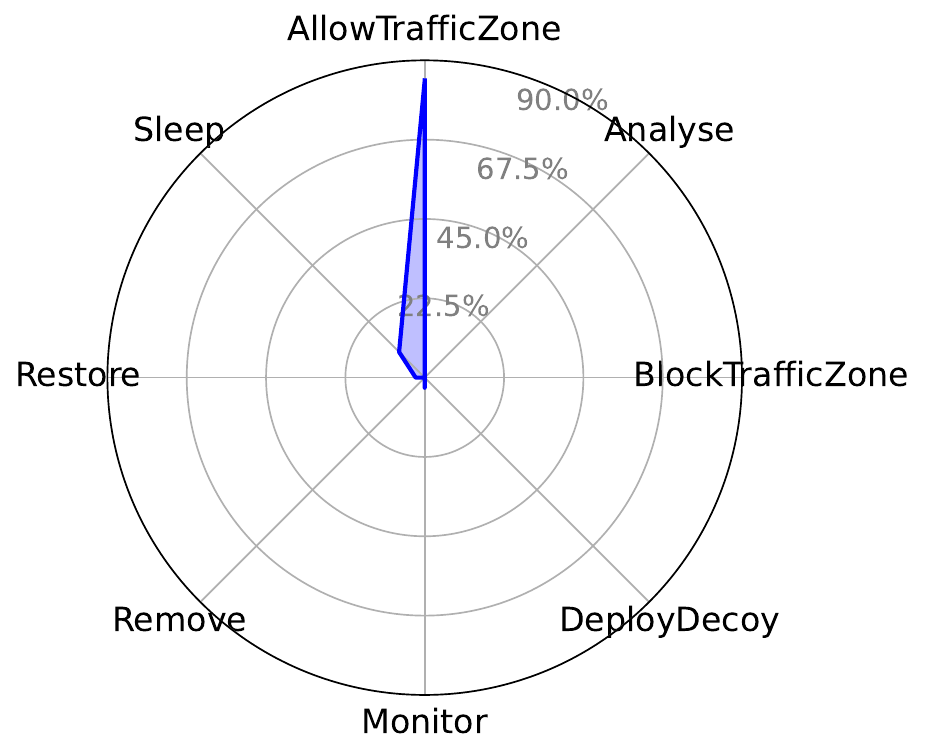}
}
\subfloat[$\bm{\pi}_{Blue}^1$]{
\label{fig:cc4:action_percentages:oz:abr1}
	\includegraphics[trim={0cm 0 0cm 0},width=0.5\columnwidth]{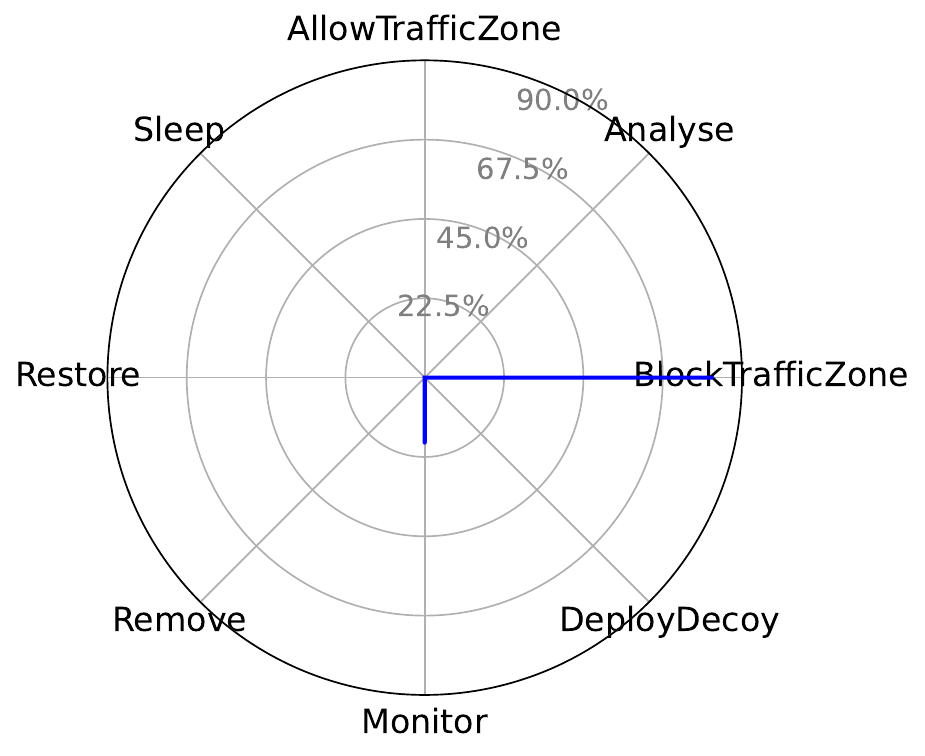}
}
}
\caption{Action percentage comparison for CC4 agents tasked with defending \emph{restricted} and \emph{operational zones A} under 
the joint-policy profiles $\langle \bm{\pi}_{Blue}^o, \bm{\pi}_{Red}^o \rangle$ and $\langle \bm{\pi}_{Blue}^1, \bm{\pi}_{Red}^o \rangle$.
Here, $\bm{\pi}_{Blue}^o$ are the original parameterisations from \cite{CC4KEEP};
$\bm{\pi}_{Red}^o$ Red's ABR against $\bm{\pi}_{Blue}^o$, and; $\bm{\pi}_{Blue}^1$, Blue's ABR against $\bm{\pi}_{Red}^o$. 
The latter agents are more active in blocking traffic, and thereby mitigating Red's attacks.   
}
\label{fig:cc4:action_percentages}
\end{figure*}


The mean Blue payoff\footnote{In CC4 each Blue agent receives an identical (team) reward.} under the initial policies is 
$\G{Blue}{\bm{\pi}_{Blue}^o, \bm{\pi}_{Red}^o} = -448.24$ (100 evaluation episodes).
The Red Vanilla and VF-PBRS responses degrade the mean payoff to $-492.33$ and 
$-484.31$ respectively.
Similarly, the GPPO oracles find better responses against 
$\bm{\pi}_{Red}^o$, $-378.27$ for Vanilla and $-368.76$ for VF-PBRS.
The value of the game following the first ABRs is 
$\G{Blue}{\mu_{Blue}, \mu_{Red}}=-408.67$.
Following five ABR iterations this value does not change significantly, 
sitting at $-412.7$.

With both Blue and Red unable to find ABRs that significantly
improve on the mixtures we switch to \say{full} 5M step responses after five ABR iterations.
However, both VF-PBRS and Vanilla Red runs converge after 2M time-steps,
and are unable to match the performance of the Red mixture agent\footnote{See \autoref{appendix:additional_results_cc4} for CC4's convergence plot and payoff matrix.}. 
%
%
The final estimated value of the game is 
$\G{Blue}{\mu_{Blue}, \mu_{Red}}=-411.91$.
%

We therefore observe a trivial change in the mixture payoffs following the first
iteration.
The reason for this becomes apparent upon inspecting the agents' action profiles.
Under the joint-policy profiles $\langle \bm{\pi}_{Blue}^o, \bm{\pi}_{Red}^o \rangle$,
agents defending the \texttt{Restricted} and \texttt{Operational} zones
of the network rarely select the \texttt{BlockTrafficZone} action 
(See \autoref{fig:cc4:action_percentages}).
Large penalties are associated with successful attacks on these areas of the network.
In contrast, $\bm{\pi}_{Blue}^1$ agents are more active in blocking traffic to these zones.
The $\bm{\pi}_{Blue}^o$ mixture weight is zero after the first ABR iteration.
 
\section{Empirical Game Augmentation Complexity} \label{sec:complexity}

While \MRO provides a means through which to conduct
a principled evaluation of cyber-defence and attacking
approaches, it does come with an increase in complexity
for augmenting the empirical payoff matrix.
Given $n = |R_{\blue}( \mA{\red})|$ approaches for computing responses for $\blue$, 
and $m = |R_{\red}( \mA{\blue})|$ approaches for $\red$,
in each iteration the MRO algorithm 
will need to compute 
$n \times |\Pi_{Red}| + m \times |\Pi_{Blue}| - n \times m$ new entries.
Here, $\Pi_{Red}$ and $\Pi_{Blue}$ represent the sets of polices available
at the start of an ABR iteration. 

A popular approach for dealing with large payoff matrices is to remove 
strategies that are found to be either strictly or weakly dominated
\citep{conitzer2005complexity,kuzmics2011elimination}\footnote{Definitions are provided in \autoref{sec:dominated}.}.
Upon evaluating the payoff matrix from CC2 after 23 ABR iterations we are unable to
identify any strictly dominated Blue policies. 
However, we do identify weakly dominated policies,
including the majority of Cardiff ABRs (excluding ABR 17) 
and GPPO ABRs 2, 7 -- 10, 12 and~18.
Therefore, an aggressive pruning would 
substantially reduce the number of evaluations required for augmenting
the empirical payoff matrix in future iterations.

Double oracle based approaches are navigating
a space of \emph{hidden game views} that is gradually expanded 
as new responses are added to the empirical game \citep{wellman2024navigating}. 
As a result policies that are currently strictly/weakly dominated 
may become relevant in future iterations.
Therefore, a periodic re-evaluation of pruned policies against recent
ABRs is likely necessary to assess their status.

\section{Discussion \& Conclusion} \label{sec:discussion}

Our empirical game-theoretic evaluations 
show the risk posed by deploying autonomous cyber-defence
agents that have only been confronted with a limited set of cyber-attack TTPs 
during training.
Once deployed, cyber-defence agents will typically be unable to make online policy adjustments for damage limitation.
Therefore, 
Blue's adversarial learning process must feature 
\red agents that pose a similar or worse threat to those that will be encountered 
once deployed.
Put plainly, even the most principled and wall-time efficient adversarial learning framework will
struggle to produce robust and resilient cyber-defence agents if the approach is 
trained against \emph{weak} Red solutions. 

Our work shows that oracles using a combination of VF-PBRS and PTMs enable 
the computation of time-efficient, robust and resilient responses. 
Our multiple response oracles algorithm provides a
principled solution, not only for
evaluating the approaches against
each other, but combining the best
policies obtained from the learning
approaches 
into a, potentially
heterogeneous, \emph{mixture agent}.
We have shown the benefits of these approaches within
the challenging context of ACD, and note their potential 
for any other application area where multiple private and public 
organisations are seeking to ensemble their custom built approaches.

\section*{Impact Statement}

Through highlighting the perils of not
viewing autonomous cyber-operations through the lens of adversarial
learning, our work is intended to serve as a wake-up call for the
ACD community.
We show that a number of design principles need to be adhered-to
for obtaining an idealised, robust, and resilient ACD agent.
These principles include:
i.) Not having to re-learn defence strategies that were previously successful against opponents;
ii.) Using a principled method for determining how to re-use past knowledge against current threats;
iii.) The ability to ensemble policies drawn from a \emph{heterogeneous set of policies};
iv.) The ability to rapidly (re)train policies, and;
v.) That learning truly robust and resilient ACD policies will only be possible if the learners are confronted with
\emph{worst-case} cyber-attacking policies.

The fifth principle naturally raises a number of ethical considerations.
Researchers and developers have justifiable concerns
that fully enabling adversarial learning could lead to the nefarious use of
cyber-attacking agents trained within a high-fidelity gym environment~\citep{cyberbattlesim}.
However, our work highlights the weaknesses of agents trained against sub-optimal opponents.
Therefore, we hope that our work will open up debates on how to confront this challenge head on,
while simultaneously ensuring compliance with relevant guidelines and best practices.

\section*{Acknowledgements}
Research funded by Frazer-Nash Consultancy Ltd. on behalf of the 
Defence Science and Technology Laboratory (Dstl) which is an 
executive agency of the UK Ministry of Defence providing world 
class expertise and delivering cutting-edge science and technology 
for the benefit of the nation and allies. 
The research supports the Autonomous Resilient Cyber Defence (ARCD) 
project within the Dstl Cyber Defence Enhancement programme.
We would also like to thank Prof. Rahul Savani (University of Liverpool) 
for valuable conversations on the topic of principled adversarial learning.

\bibliography{main}
\bibliographystyle{icml2025}

\appendix

\section{Theoretical Analysis} \label{sec:theory}

\subsection{Approximate Double Oracles} \label{sec:adoproof}

In games that suffer from the curse-of-dimensionality 
an oracle can at best hope to find an \emph{approximate best response} (ABR)~\citep{oliehoek2018beyond}:
%
%
%
\begin{definitionsec}[Approximate Best Response]\label{def:ABR}
A policy $\pi_i \in \Pi_i$ of player $i$ is an 
ABR, 
also referred to as a resource bounded best response (RBBR)~\citep{oliehoek2018beyond}, 
against a \mop $\mu_j$, \emph{iff},
\begin{equation}
\forall \pi'_i \in \Pi_i, 
\G{i}{\JointPolicy{\pi_i}{\mu_j}}
\geq 
\G{i}{\JointPolicy{\pi'_i}{\mu_j}}.
\end{equation}
\end{definitionsec}
ABRs are also used to estimate the exploitability $\Exploitability$ of the current mixtures:
\begin{equation}
\Exploitability \gets
\G{i}{\JointPolicy{\O{i}{\mu_j}}{\mu_j}} + 
\G{j}{\JointPolicy{\mu_i}{\O{j}{\mu_i}}}
\end{equation}
If $\Exploitability \leq 0$, then the oracles have failed to find 
ABRs, 
and a \emph{resource bounded Nash equilibrium} (RBNE)
has been found~\citep{oliehoek2018beyond}.
Resources in this context refers to the amount of computational power 
available for obtaining an ABR:
\begin{definitionsec}[Resource Bounded Nash Equilibrium] \label{def:RBNE}
Two mixtures of policies $\JointPolicy{\mu_1}{\mu_2}$ are a resource-bounded Nash equilibrium \emph{iff},
\begin{equation} \label{eq:termination_condition}
\forall_i \G{i}{\JointPolicy{\mu_i}{\mu_j}}
\geq
\G{i}{\JointPolicy{\O{i}{\mu_j}}{\mu_j}}.
\end{equation}
\end{definitionsec}

Extending a proof from~\citep{oliehoek2018beyond} for generative adversarial networks (GANs)~\citep{goodfellow2014generative},
we can show that $\Exploitability \leq 0$ implies we 
have found a resource bounded Nash equilibrium for the \ADO algorithm when applied to adversarial deep reinforcement learning: 
\begin{theoremsec}
If the termination condition from Equation  \ref{eq:termination_condition} is met, then a resource bounded Nash equilibrium has been found.
\end{theoremsec}

\begin{proof}
It can be shown that $\Exploitability \leq 0$ 
implies that the agents have converged upon a resource bounded Nash equilibrium:
\begin{equation}
\begin{aligned} 
\Exploitability &= 
\G{i}{\JointPolicy{\O{i}{\mu_j})}{\mu_j}} +  
\G{j}{\JointPolicy{\mu_i}{\O{j}{\mu_i}}} \\ 
& \leq 0 = \G{i}{\JointPolicy{\mu_i}{\mu_j}} + 
\G{j}{\JointPolicy{\mu_i}{\mu_j}}.
\end{aligned} 
\end{equation}
Note that, as per Definition \ref{def:ABR},
\begin{equation} \label{eq:abr_property}
\G{i}{\JointPolicy{\O{i}{\mu_j}}{\mu_j}} \geq
\G{i}{\JointPolicy{\mu_i}{\mu_j}}
\end{equation}
for all computable $\pi_i \in \Pi^{O}_i$, where $\Pi^{O}_i$ is the set of ABRs that can be computed using an oracle $O$ for player~$i$. 
The same holds for player~$j$.
Therefore, the only way that Equation  \ref{eq:abr_property} could fail to hold 
is, if $\mu_i$ includes policies that are not 
computable (not in $\Pi_i^{O}$) that provide a higher payoff. 
However, the support of $\mu_i$ is composed of policies computed in previous iterations, therefore this cannot be the case. 
%
%
%
%
Together with Equation \ref{eq:abr_property} this directly implies the following for each agent: 
\begin{equation}
\G{i}{\JointPolicy{\mu_i}{\mu_j}} = 
\G{i}{\JointPolicy{\O{i}{\mu_j}}{\mu_j}}. 
\end{equation}
We have therefore found a resource bounded Nash equilibrium.
\end{proof}

\begin{theoremsec}
Given sufficient resources to compute $\epsilon$-best responses, 
then a resource bounded Nash equilibrium is an $\epsilon$-Nash 
equilibrium ($\epsilon$-NE)~\citep{oliehoek2018beyond}.
\end{theoremsec}

\begin{proof} \label{proof:RBNE_epsNE}
Starting from the resource bounded Nash equilibrium with respect to the mixture profiles $\langle \mu_i, \mu_j \rangle$, we shall assume an arbitrary 
response by agent $i$.
By definition of a resource bounded Nash equilibrium (Definition \ref{def:RBNE}): 
\begin{multline}
\G{i}{\JointPolicy{\mu_i}{\mu_j}}
\geq
\G{i}{\JointPolicy{\O{i}{\mu_j}}{\mu_j}}
\geq \\
\max_{\mu'_i}
\G{i}{\JointPolicy{\mu'_i}{\mu_j}} - \epsilon.
\end{multline}
\end{proof}

\subsection{Multiple Response Oracles} \label{sec:multi_oracle:theory} 

\subsubsection{Convergence Guarantees} \label{sec:multi_oracle:theory:convergence} 

In this section we shall first theoretically show that 
our \MRO algorithm does not disrupt the convergence
guarantees of the double oracle formulation.
Next, we shall show that, given a set of assumptions,
\MRO is guaranteed to produce stronger mixture 
agents than the \ADO.

We begin with the termination condition, where we note 
that $\Exploitability$ can now be computed with respect 
to the best performing approaches from each agent's 
respective sets of responses:
%
\begin{multline} \label{eq:termination_condition_mro}
\Exploitability \gets
 \G{i}{\JointPolicy{O_i(R_i(\mA{j}))}{\mu_j}} + \\
 \G{j}{\JointPolicy{\mu_i}{O_j(R_j(\mA{i}))}}.
\end{multline}
%
In the above equation, 
$O_i(R_i(\mA{j}))$ and $O_j(R_j(\mA{i}))$ return $\pi^*_{i}$ and $\pi^*_{j}$, 
\ie the best response within
their respective response sets 
$\pi^*_{i} \in R_i(\mu_j)$ and 
$\pi^*_{j} \in R_j(\mu_i)$, 
as defined in \autoref{eq:mro:abr_policies}.
%
%
Therefore, it remains the case that if $\Exploitability \leq 0$, 
then the oracles have failed to improve on the current mixtures, 
and a \emph{resource bounded Nash equilibrium} (RBNE) has been found.


\begin{proof}
It can be shown that $\Exploitability \leq 0$ 
implies that the agents have converged upon a resource bounded Nash equilibrium for the \MRO algorithm:
\begin{equation}
\begin{aligned} 
\Exploitability & = \G{i}{\JointPolicy{O_i(R_i(\mA{j}))}{\mu_j}} +  
\G{j}{\JointPolicy{\mu_i}{O_j(R_j(\mA{i}))}} \\
& \leq 0 = \G{i}{\JointPolicy{\mu_i}{\mu_j}} + 
\G{j}{\JointPolicy{\mu_i}{\mu_j}}.
\end{aligned} 
\end{equation}
As per Definition \ref{def:ABR} and \autoref{eq:mro:abr_policies}, we have:
\begin{equation} \label{eq:abr_property2}
\G{i}{\JointPolicy{\O{i}{R_i(\mu_j)}}{\mu_j}} \geq
\G{i}{\JointPolicy{\mu_i}{\mu_j}}
\end{equation}
for all computable $\pi_i \in \Pi^{O}_i$, where $\Pi^{O}_i$ is the set of ABRs 
that can be computed using an oracle $O$ for player~$i$. 
Therefore, given that $\pi^{*}_{i} \gets O_i(R_i(\mA{j}))$, 
\begin{equation} \label{eq:abr_property3}
\G{i}{\JointPolicy{\pi^*_{i}}{\mu_j}} \geq
\G{i}{\JointPolicy{\mu_i}{\mu_j}}.
\end{equation}
The same holds for player~$j$.
The only way that \autoref{eq:abr_property3} could fail to hold 
is if $\mu_i$ includes policies that are not 
computable by $R_i$, and therefore not in $\Pi_i^{O}$, that provide a higher payoff. 
However, the support of $\mu_i$ is composed of policies computed in previous iterations, therefore this cannot be the case. 
We conclude that 
\begin{equation}
\G{i}{\JointPolicy{O_i(R_i(\mA{j}))}{\mu_j}} \geq 
\G{i}{\JointPolicy{\mu_i}{\mu_j}}
\end{equation}
for each agent, based on the respective set of responses computed using the resource bounded
response function~$R_i$.
Together with Equation \ref{eq:abr_property2} this directly implies the following for each agent: 
\begin{equation}
\G{i}{\JointPolicy{\mu_i}{\mu_j}} = 
\G{i}{\JointPolicy{O_i(R_i(\mA{j}))}{\mu_j}}. 
\end{equation}
We have therefore found a resource bounded Nash equilibrium.
\end{proof}

\subsubsection{Improved Mixtures} \label{sec:multi_oracle:theory:improved_mixtures} 

Next we show that, based on the following assumptions, our \MRO algorithm 
is guaranteed to produce stronger mixtures compared to the \ADO algorithm.
\begin{assumption}\label{as:1}
There exists a policy $\pi^{i,n}_{p_1}$ in the support for player $p_1$,
that therefore has a weighting $w_i > 0$ within the mixture $\mu_{p_1}$.
However, $\pi^{i,n}_{p_1}$ was not the ABR against $\mu^{n}_{p_2}$. 
Here, $n$ represents the best response
iteration during which $\pi^{i,n}_{p_1}$ was
computed.
Formally, there exists another policy $\pi^{j,n}_{p_1}$ that was a better response
to the opponent mixture in iteration $n$:
\begin{equation}
\exists \pi^{j,n}_{p_1}, \mathcal{G}(\langle \pi^{j,n}_{p_1}, \mu^n_{p_2} \rangle) > \mathcal{G}(\langle \pi^{i,n}_{p_1}, \mu^{n}_{p_2} \rangle). 
\end{equation}
\end{assumption}
\begin{assumption}\label{as:2}
The policy $\pi^i_{p_1}$ belongs either to a \emph{dominant strategy equilibrium}
or a \emph{mixed strategy equilibrium} that has undergone a process of 
iteratively eliminating weakly dominated policies using the empirical payoff matrix~$\mathcal{N}$.
\end{assumption}
Therefore, given the above assumptions, 
the policy $\pi^i_{p_1}$ cannot be removed from the agent's set of policies 
without reducing the gain for $p_1$, giving us a minimal viable support:
\begin{definitionsec}[Minimal Viable Support] \label{def:MMS}
Given a mixture $\mu_{p_1}$ with a minimum viable support set of polices, we cannot
remove a policy $\pi^{i,n}_{p_1}$ without reducing the gain for $p_1$. 
\begin{equation} \label{eq:policy_contribution}
\mathcal{G}_{p_1}(\langle \mu_{p_1}, \mu_{p_2} \rangle) > \mathcal{G}_{p_1}(\langle \mu_{p_1}', \mu_{p_2} \rangle),
\end{equation}
%
with $\mu_{p_1}'$ representing a mixture profile without $\pi^{i,n}_{p_1}$. 
\end{definitionsec}

\begin{theoremsec} \label{def:impact_of_removing_policy}
As per equation \autoref{eq:policy_contribution}, 
we cannot remove the policy $\pi^{i,n}_{p_1}$ from $\mu_{p_1}$ 
without impacting the value of the game.
Therefore, by adding $\pi^{i,n}_{p_1}$ to the mixture $\mu_{p_1}$, 
the \MRO algorithm is able to obtain a stronger mixture policy for $p_1$ 
than the traditional \ADO algorithm with a mixture 
$\mu_{p_1}'$.
\end{theoremsec}

%
\begin{proof}
\ADO's mixture $\mu_{p_1}'$ outperforming the \MRO algorithm's $\mu_{p_1}$ would imply that:
\begin{equation} \label{eq:contradiction}
\mathcal{G}_{p_1}(\langle \mu_{p_1}', \mu_{p_2} \rangle) > \mathcal{G}_{p_1}(\langle \mu_{p_1}, \mu_{p_2} \rangle).
\end{equation}
However, as per Definition \ref{def:impact_of_removing_policy}, 
we cannot remove $\pi^{i,n}_{p_1}$ without reducing the gain for $p_1$.
Therefore, \autoref{eq:contradiction} could only be true, iff, $\mu_{p_1}'$ contained
additional policies not found within $\mu_{p_1}$.
However,
\begin{equation}
\mu_{p_1}' \subset \mu_{p_1}.
\end{equation}
Therefore, the condition from \autoref{eq:policy_contribution} applies, 
meaning the \MRO algorithm has found a mixture that achieves a larger gain when compared
to the mixture obtained by the \ADO algorithm.
\end{proof}

%

\subsection{VF-PBRS - Proof of Sufficiency} \label{app:proof_of_sufficiency}

%
\cite{ng1999policy} show that 
a reward shaping function $F$ being a \emph{potential-based shaping function} 
is a necessary and sufficient condition to guarantee consistency when learning
an optimal policy on an MDP  
$\M' = (\states, \actions, \transition, \rewards + F, \gamma)$
instead of 
$\M = (\states, \actions, \transition, \rewards, \gamma)$. 

Regarding necessity, the authors note that if $F$ is not a potential-based shaping function,
then there exist transition and reward functions such that no optimal policy in $\M'$ is 
also optimal in $\M$. 
With respect to sufficiency, if $F$ is a potential-based shaping function,
every optimal policy in $\M'$ will also be optimal in~$\M$.

For convenience we include \cite{ng1999policy}'s proof of sufficiency below.
As in the original PBRS paper the proof assumes scenario-optimal policies
can be obtained, hence we relate it to VF-PBRS within the context
of double oracles rather than the \ADO algorithm, 
where at best we can hope to obtain \emph{approximate} best responses. 
First, to recap:

\begin{theoremsec} \label{therome:sufficiency}
We assume a state space $\states$, actions $\actions$ and discount factor $\gamma$.
Given a reward shaping function $F : \states \times \actions \times \states \rightarrow \realnumber$,
$F$ is considered a \emph{potential-based shaping function} if there 
exists a real-value function $\Phi : \states \rightarrow \realnumber$ such that for all 
$s \in S - \{s_0\}, a \in \actions, s' \in \states$, where $s_0$ represents
an absorbing state,
\begin{equation}
F(s,a,s') = \gamma \Phi(s') - \phi(s).
\end{equation}
Note that in the above equation, 
if $\states$ does not have an absorbing state,
$\states - \{s_0\} = \states$ then $\gamma < 1$.
Given the above, $F$ being a potential-based shaping function is a sufficient 
condition to guarantee that any policy obtained under $\M' = (\states, \actions, \transition, \rewards + F, \gamma)$
will be consistent with an optimal policy for $\M = (\states, \actions, \transition, \rewards, \gamma)$ (and vice versa). 
\end{theoremsec}
\begin{proof}
We assume a VF-PBRS function $F$ of the form provided in \autoref{eq:pbrs:ensemble}. 
If $\gamma = 1$, the replacing of $\Phi(s)$ with $\Phi'(s) = \Phi(s) - k$
for any constant $k$ will not impact the shaping reward $F$.
This is due to $F$ being the difference in potentials.
Therefore, given an absorbing state $s_0$, we can 
replace $\Phi(s)$ with $\Phi'(s) = \Phi(s) - \Phi(s_0)$, 
and assume that our potential function $\Phi(s)$ satisfies $\Phi(s_0) = 0$.

We know that for the original MDP $\M$ the optimal Q-function satisfies 
the Bellman Equations~\citep{sutton2018reinforcement}:
\begin{multline}
Q^*_\M(s,a) = \E_{s'~P_{s,a}(\cdot)}\big[\rewards(s,a,s') +\\ \gamma \max_{a' \in \actions}Q^*_\M(s', a') \big].
\end{multline}
If we subtract $\Phi(s)$ from both sides and apply some simple
algebraic manipulations we get:
\begin{multline} \label{eq:pbrs_proof1.1}
Q^*_\M(s,a)  - \Phi(s) = \E_{s'~P_{s,a}(\cdot)}\big[\rewards(s,a,s') + \\ 
\gamma \Phi(s') - \Phi(s) + \\ 
\gamma \max_{a' \in \actions}\big(Q^*_\M(s', a') - \Phi(s')\big) \big].
\end{multline}
Upon defining $\hat{Q}_{\M'}(s, a) \triangleq Q^*_\M(s,a)  - \Phi(s)$ and substituting that and $F(s, a, s') = \gamma \Phi(s') - \Phi(s)$ into \autoref{eq:pbrs_proof1.1}, we get:
\begin{equation} \label{eq:pbrs_proof1.2}
\begin{split}
&\hat{Q}_{\M'}(s, a) \\
& = \E_{s'}\big[\rewards(s,a,s') + F(s, a, s') + \gamma \max_{a' \in \actions}\hat{Q}_{\M'}(s', a') \big],\\
& = \E_{s'}\big[\rewards'(s,a,s') + \gamma \max_{a' \in \actions}\hat{Q}_{\M'}(s', a') \big].
\end{split}
\end{equation} 
This is the Bellman equation for $\M'$. Therefore:
\begin{equation}
Q^*_{\M'}(s, a) = \hat{Q}_{\M'}(s, a) = Q^*_{\M}(s, a) - \Phi(s).
\end{equation}
As a result, the optimal policy found by our oracle using VF-PBRS for $M'$ satisfies:
\begin{equation} \label{eq:pbrs_proof1.3}
\begin{split}
\pi^*_{\M'}(s) & = \argmax_{a \in \actions} Q^*_{\M'}(s,a), \\
               & = \argmax_{a \in \actions} Q^*_{\M}(s,a) - \Phi(s), \\
               & = \argmax_{a \in \actions} Q^*_{\M}(s,a).
\end{split}
\end{equation} 
\end{proof}

%
%
%

\section{Network Diagrams} \label{app:network_diagram}

\subsection{CAGE Challenge 2}

\begin{figure}[H]
\includegraphics[width=1.0\columnwidth]{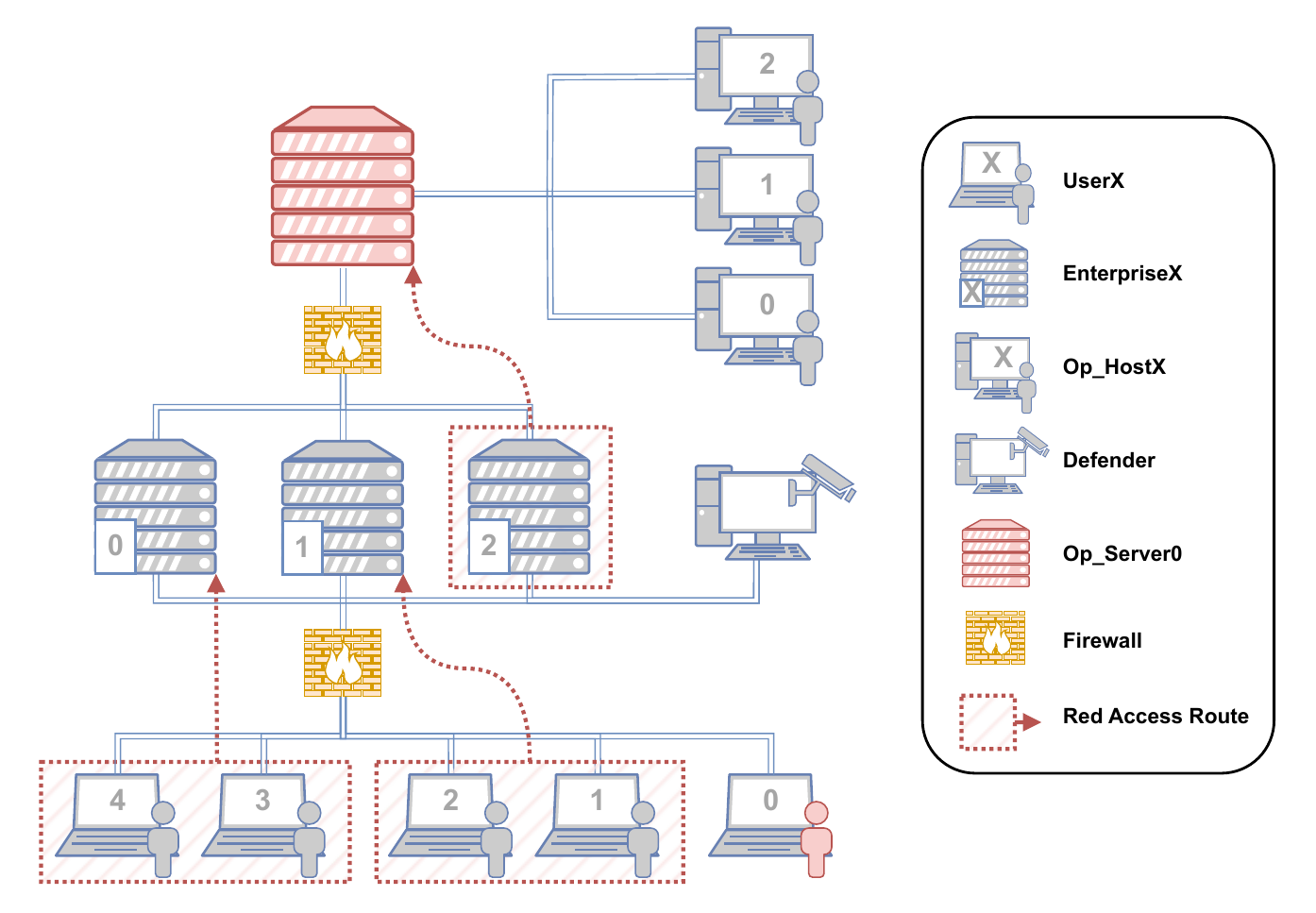}
\caption{CAGE Challenge 2 Network Layout.}
\label{fig:cc2_network_diagram}
\end{figure}

In \autoref{fig:cc2_network_diagram} we provide an illustration of the network layout for CC2.
We train our agents on \texttt{Scenario2}, and include a depiction of the
Red agent's access routes (red dotted line). 
A detailed description of this challenge can be found in \cite{cage_challenge_2_announcement}.

\subsection{CAGE Challenge 4}

\begin{figure}[H]
\includegraphics[width=1.0\columnwidth]{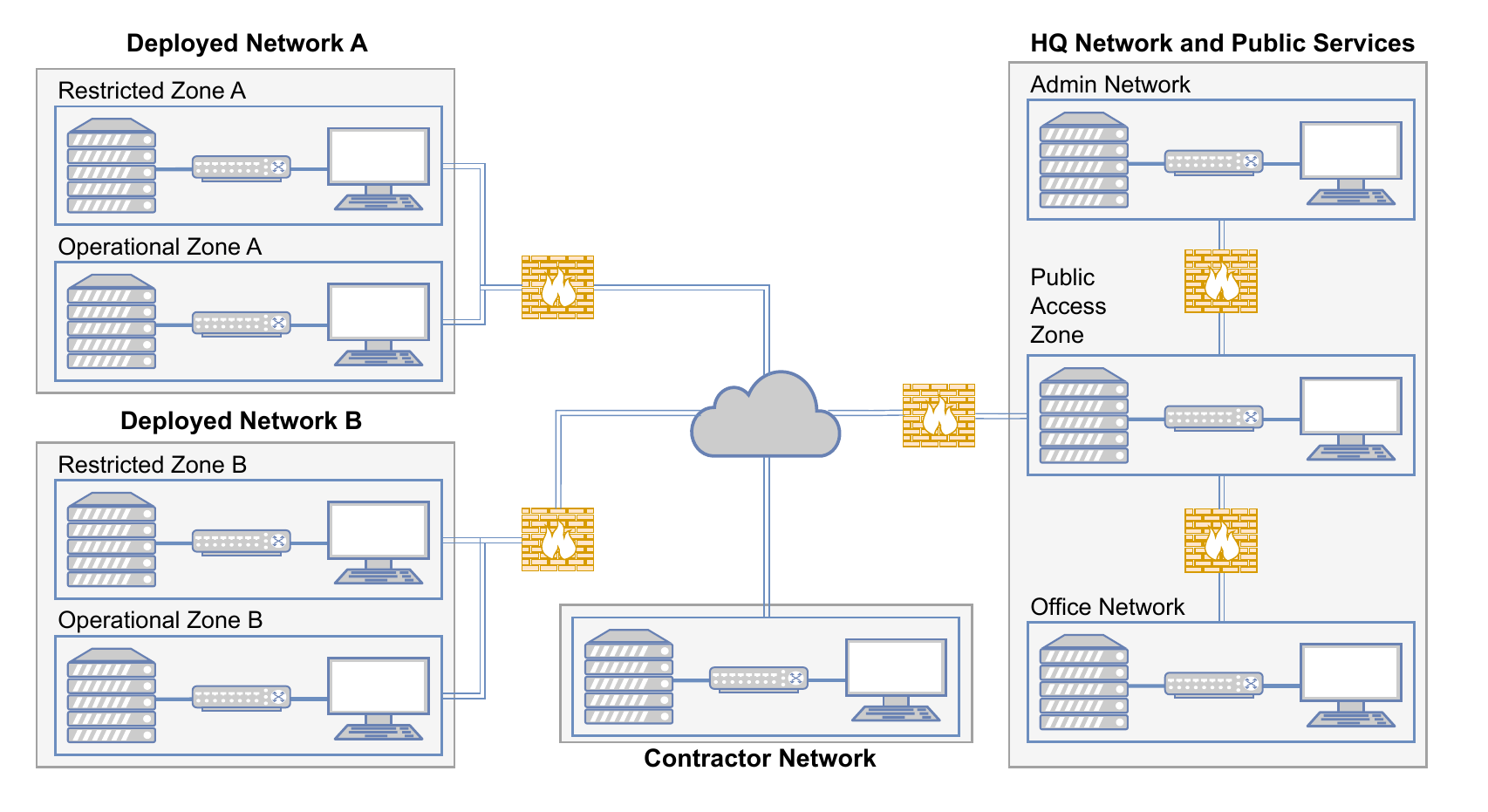}
\caption{CAGE Challenge 4 Network Layout.}
\label{fig:cc4_network_diagram}
\end{figure}

\autoref{fig:cc4_network_diagram} depicts the network layout for CC4. 
The network is defended by five Blue agents. 
Deployed networks A and B have two ACD agents each, one per Restricted / Operational zone.
The remaining ACD agent is tasked with defending the HQ network. 
At every step there is a small probability that a Red cyber-attacking agent will spawn 
if a Green (user) agent opens a phishing email. 
Red agents can also spawn upon a Green user accessing a compromised service.
Further details can be found in \cite{cage_challenge_4_announcement}.

\section{Experiment Settings} \label{app:experiment_settings}

Below we provide an overview of our experiment settings.
We begin by outlining the configurations that are used across our MRO runs.
We subsequently provide details regarding CAGE Challenge 2 and~4 specific configurations. 

\textbf{Policy evaluations:} Policy evaluations consist of $100$ evaluation episodes. 
This includes: i.) The evaluation 
of responses following each ABR iteration; 
ii.) Payoff table augmentations; 
iii.) Significance testing, and; 
iv.) The visualisation of policy characteristics.

\textbf{Pre-Trained Model Sampling:} With respect to PTM sampling, both our CC2 and CC4 runs use our discussed 
$\epsilon$-greedy based approach for gradually sampling from policies computed during
the respective runs. An $\epsilon$ discount rate of 0.95 is applied following each ABR iteration.

\textbf{VF-PBRS:}
Unless specified otherwise, the VF-PBRS oracles use a scaling coefficient $\tau=1$. 
A further key parameter for VF-PBRS is the value to select for $\gamma$.
As noted by \cite{ng1999policy}:
\say{Future experience with potential-style shaping rewards may also
lead one to occasionally try shaping rewards that are inspired by potentials,
but which are perhaps not strictly of the form given}. 
The authors go on to note that for certain problems, it may be easier for 
an expert to propose an undiscounted shaping function $\Phi(s') - \Phi(s)$,
even when using $\gamma \neq 1$ for training the agents.
%

\cite{grzes2009theoretical} note that when given two states $s$ and $s'$, 
such that $\Phi(s) < \Phi(s')$, then the agent should be rewarded by the 
shaping reward for transitioning from $s \rightarrow s'$.
Furthermore, the agent should not be rewarded for transitioning from 
$s' \rightarrow s$ or remaining in $s$, $s \rightarrow s$.
Therefore, given a potential function $\Phi$, 
a suitable value for $\gamma$ is required where the following properties hold
for the mentioned $s$ and $s'$:
\begin{equation} \label{eq:phi_pos}
F(s, s') = \gamma \Phi(s') - \Phi(s) \geq 0,
\end{equation}
\begin{equation} \label{eq:phi_neg}
F(s', s) = \gamma \Phi(s) - \Phi(s') \leq 0,
\end{equation}
\begin{equation} \label{eq:phi_eq}
F(s, s) = \gamma \Phi(s) - \Phi(s) \leq 0.
\end{equation}

\cite{grzes2009theoretical} show that finding a $\gamma$ that meets 
the above requirements is challenging, 
even when dealing with well defined heuristics, such as the straight line 
distance to a given target.
In this work we observe that this challenge is amplified when utilising
value functions as potential functions.
Therefore, we consider two discount rates,  using $\gammapbrs$ to refer to 
the discount factor applied to the potential of
the follow-on state $\Phi(s')$. 
We often encounter scenarios where, 
despite $\Phi(s) < \Phi(s')$, we  observe  $\gammapbrs \Phi(s') - \Phi(s) < 0$
when using $\gammapbrs = \gamma$ with a standard discount rate, e.g.,
$0.99$. 
As a result, for our current experiment we use a discount rate $\gamma=1$.
Evaluating the impact of the value of $\gamma$ on VF-PBRS
for double oracles represents and interesting avenue for future research.
 
\subsection{Cage Challenge 2}
 
For CAGE Challenge 2 we use the default configuration 
provided with the repository for conducting our training
runs. 
Each training and evaluation episode consisted of 
$100$ time-steps.
Unless specified otherwise, each oracle uses 1.5M
training steps per response. 
Below we provide an overview of oracle specific configurations.

\textbf{Cardiff University (Blue):}
We make the following changes in order to utilise the CAGE Challenge 2 winning approach
from \cite{JohnHannay} as an oracle:
i.) We remove the fingerprinting mechanism designed to identify the rules-based Red agents 
used in the challenge, and;
ii.) We add parallel data gathering to reduce wall-time~\citep{schulman2017proximal}.
As a starting point, we subsequently obtain a \texttt{CardiffUniform} agent, 
which is trained in a setting where we uniformly sample over the two provided rules-based \red agents (B-line and Meander).
\texttt{CardiffUniform} achieves comparable results to the 
version using fingerprinting ($-14.74$ compared to $13.76$  
against the Meander, and $-16.85$ compared to $16.6$ against
B-line, with the latter scores being those achieved by 
Cardiff's submission to CAGE Challenge 2).

While we find all \texttt{CardiffUniform} (and provided)
Cardiff university approaches to be highly exploitable
when confronted with Red ABRs,
we note that the approach has a lot of merit 
with respect to: i.) Reducing the action space in relation 
to available decoys, using domain knowledge, and; ii.)  
Buffers for keeping track of various properties throughout each episode. 
Therefore, we hypothesise that the Cardiff implementation can 
be used to reduce exploitability, \emph{if} trained within an 
adversarial learning framework where the approach is presented 
with a set of more challenging \red opponents.
For this purpose we add the Cardiff approach as a response
oracle.

With respect to hyperparameters our oracles use the same ones
as those provided by \cite{JohnHannay}:
\begin{table}[H]
\centering
\resizebox{\columnwidth}{!}{
\begin{tabular}{c|c}
\textbf{Hyperparameter} & \textbf{Value} \\
\hline
Optimiser: & Adam~\citep{kingma2014adam} \\
Learning rate: & $0.002$ \\
VF loss coefficient: & $0.5$ \\
Adam betas: & $0.9$, $0.99$ \\
Clipping $\epsilon$: & $0.2$ \\
Discount rate $\gamma$: & $0.99$ \\
Num. epochs: & $6$ \\
Num. workers: & $10$ \\
Horizon (Environment Timesteps): & $100$ \\
Entropy coefficient: & $0.01$ \\
Batch Size: & $1,000$ \\
\end{tabular}}
\caption{Hyperparameters used by the Cardiff oracle for CC2.}
\label{tab:hyperparams:cardiff}
\end{table}

\textbf{Cybermonic Graph Inductive PPO (Blue):}
No modifications are made to \cite{CC2KEEP}'s Graph-PPO agents.
The hyperparameters used by the algorithm are: 
\begin{table}[H]
\centering
\resizebox{\columnwidth}{!}{
\begin{tabular}{c|c}
\textbf{Hyperparameter} & \textbf{Value} \\
\hline
Optimiser: & Adam~\citep{kingma2014adam} \\
Actor learning rate: & $0.0003$ \\
Critic learning rate: & $0.001$ \\
VF loss coefficient: & $0.5$ \\
Clipping $\epsilon$: & $0.2$ \\
Discount rate $\gamma$: & $0.99$ \\
Num. epochs: & $4$ \\
Num. workers: & $10$ \\
Horizon (Environment Timesteps): & $100$ \\
Entropy coefficient: & $0.01$ \\
Batch Size: & $2,048$ \\
\end{tabular}}
\caption{Inductive GPPO Hyperparameters.}
\label{tab:hyperparams:cybermnic}
\end{table}

\textbf{Action-Masking PPO (Red):}

Our action masking PPO agent is implemented within RLlib,
and masks invalid actions through adding a large negative 
value to the respective action logits.
The hyperparameters used by the algorithm are: 
\begin{table}[H]
\centering
\resizebox{\columnwidth}{!}{
\begin{tabular}{c|c}
\textbf{Hyperparameter} & \textbf{Value} \\
\hline
Optimiser: & Adam~\citep{kingma2014adam} \\
Learning rate: & $1e-5$ \\
VF loss coefficient: & $0.25$ \\
Clipping $\epsilon$: & $0.2$ \\
Discount rate $\gamma$: & $0.99$ \\
Num. epochs: & $4$ \\
Num. workers: & $10$ \\
Horizon (Environment Timesteps): & $100$ \\
Entropy coefficient: & $0.005$ \\
Batch Size: & $1,000$ \\
\end{tabular}}
\caption{Action-Masking PPO Hyperparameters.}
\label{tab:hyperparams:amppo}
\end{table}

\subsection{Cage Challenge 4}

\textbf{Cybermonic Graph Inductive Multi-Agent PPO (Blue):}
The main differences from the single agent Graph-PPO agent framework by \cite{CC2KEEP}
and the multi-agent version \citep{CC4KEEP} are as follows:
\begin{table}[H]
\centering
\begin{tabular}{c|c}
\textbf{Hyperparameter} & \textbf{Value} \\
\hline
Num. workers: & $25$ \\
Horizon (Environment Timesteps): & $500$ \\
Batch Size: & $2,500$ \\
\end{tabular}
\caption{Multi-Agent GPPO Hyperparameters.}
\label{tab:hyperparams:cybermnic}
\end{table}

\textbf{Action-Masking Multi-Agent PPO (Red):}
The hyperparameters used by our MAPPO Red agent
are largely unchanged, compared to the single 
agent case.
The only differences being:
\begin{table}[H]
\centering
\begin{tabular}{c|c}
\textbf{Hyperparameter} & \textbf{Value} \\
\hline
Num. workers: & $25$ \\
Horizon (Environment Timesteps): & $500$ \\
Entropy coefficient: & $0.001$ \\
Batch Size: & $1,024$ \\
\end{tabular}
\caption{Action-Masking MAPPO Hyperparameters.}
\label{tab:hyperparams:ammappo}
\end{table}

\section{Strictly and Weakly Dominated Strategies} \label{sec:dominated}

In this section we shall consider the different types of mixtures 
$\langle \mu_{i}, \mu_{j} \rangle$ that a Nash solver may
yield for agents $i$ and $j$ given a normal-form game $\mathcal{N}$,
and the definitions for strictly and weakly dominated strategies.
First we distinguish between a pure and a mixed strategy equilibrium.
As the name indicates, for a pure strategy Nash equilibrium both players
place a 100\% probability on playing a single strategy respectively.
A mixed strategy Nash equilibrium, meanwhile, involves at least one player
using a weighted randomised strategy selection process.
\cite{von2007theory} showed that a mixed-strategy Nash equilibrium will exist 
for any zero-sum game with a finite set of strategies.

Here, it is worth noting that, given a mixed strategy Nash equilibrium 
$\langle \mu_{i}, \mu_{j} \rangle$, for a mixed strategy $\mu_{i}$
each pure strategy included in the mix (\ie with a weighting above zero)
is itself a best response against $\mu_{j}$, and will yield the same
payoff as $\mathcal{G}_i(\langle \mu_{i}, \mu_{j} \rangle)$~\citep{fang2021introduction}.
\begin{theoremsec}[Mixed Strategy Theorem] \label{theorem:minmax_strategy_theorem} 
If a mixed strategy $\mu_{i}$, belonging to player $i$, is a best
response to the (mixed) strategy $\mu_{j}$ of player $j$, then,
for each pure strategy (action) $\action_k$ with a weight $\mu_{i}(\action_k) > 0$
it must be the case that $\action_k$ is itself a best response to $\mu_{j}$.
Therefore, $\mathcal{G}_i(\langle \action_{k}, \mu_{j} \rangle)$ must be the same
for all strategies (actions) included in the mix.
\end{theoremsec}

We note that a payoff matrix may feature \emph{strictly dominated strategies},
\emph{weakly dominated strategies} and \emph{strictly dominant strategies}.
Strictly dominated strategies (actions) are strategies that always deliver a worse outcome 
than at least one alternative strategy, regardless of the strategy chosen by the 
opponent. 
Strictly dominated strategies can, therefore, be safely deleted from a normal-form game. 
We provide formal definitions for the remaining concepts below~\citep{slantchev2008game}.  
\begin{definitionsec}[Strictly Dominant Action]
Strictly dominant actions are always optimal, therefore, $a_i^* \in \actions$ is
a \textbf{strictly dominant action} for player $i$ iff:
$$
\G{i}{\JointPolicy{a^*_{i}}{a_{-i}}} > \G{i}{\JointPolicy{a_{i}}{a_{-i}}} \forall a_i \neq a^*_i, \forall a_{-i} \in \actions_{-i}.
$$
\end{definitionsec}
Within the above definition $\actions_{-i}$ refers to the set of opponent actions. 
Next we consider weakly dominated actions:
\begin{definitionsec}[Weakly Dominant Action]
An action is weakly dominant action if it is always optimal, while every other action is sometimes sub-optimal:
$$
\G{i}{\JointPolicy{a^*_{i}}{a_{-i}}} \geq \G{i}{\JointPolicy{a_{i}}{a_{-i}}} \forall a_i \neq a^*_i, \forall a_{-i} \in \actions_{-i},
$$
%
with a strict inequality for some $a_{-i} \in \actions_{-i}$, for any~$a_i \neq a^*_i$. 
\end{definitionsec}

It is often feasible, within zero-sum games, to remove 
weakly dominated strategies without impacting the value 
of the game.
However, the order in which weakly dominated strategies are eliminated 
can impact the value of the subsequent \emph{sub-game}, 
compared to original game.
Determining the correct order for removing weakly dominated strategies from a normal-form game,
without impacting the value of the game, 
represents a computationally challenging problem, 
with respect to the number of strategies available~\citep{brandt2011complexity}.

\section{Payoff Matrix Augmentation} \label{app:payoff_augmentation_complexity_vis}

\begin{figure}[H]
\centering
\resizebox{\columnwidth}{!}{
\subfloat[Payoff Matrix Growth]{
\label{fig:payoff_table_growth}
\includegraphics[width=0.5\columnwidth]{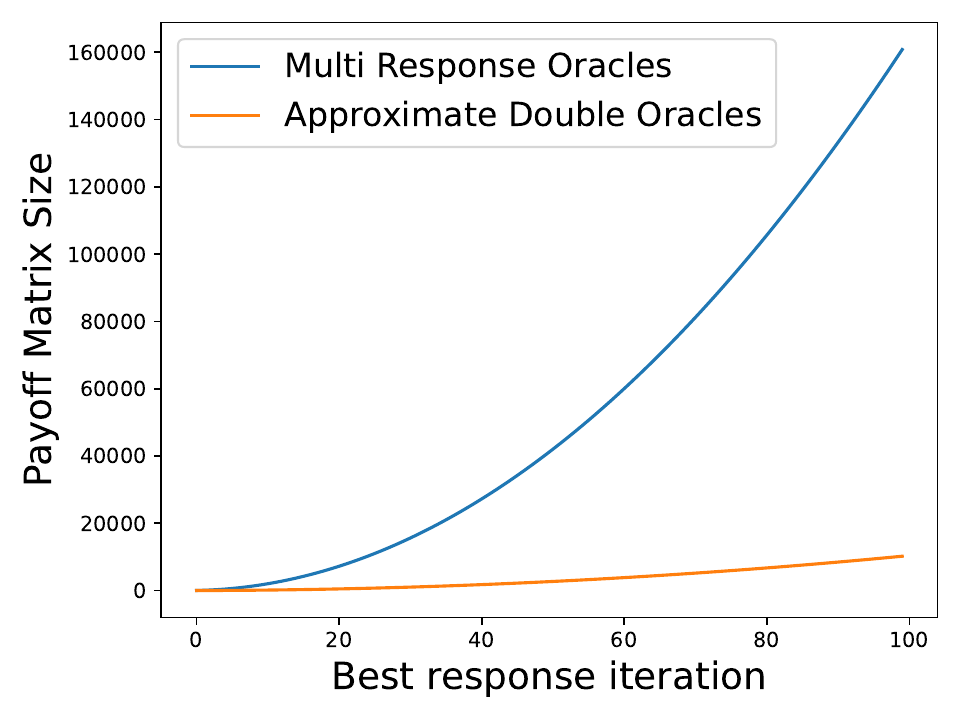}}
\subfloat[Matrix Augmentations]{
\label{fig:number_of_augmentations}
\includegraphics[width=0.5\columnwidth]{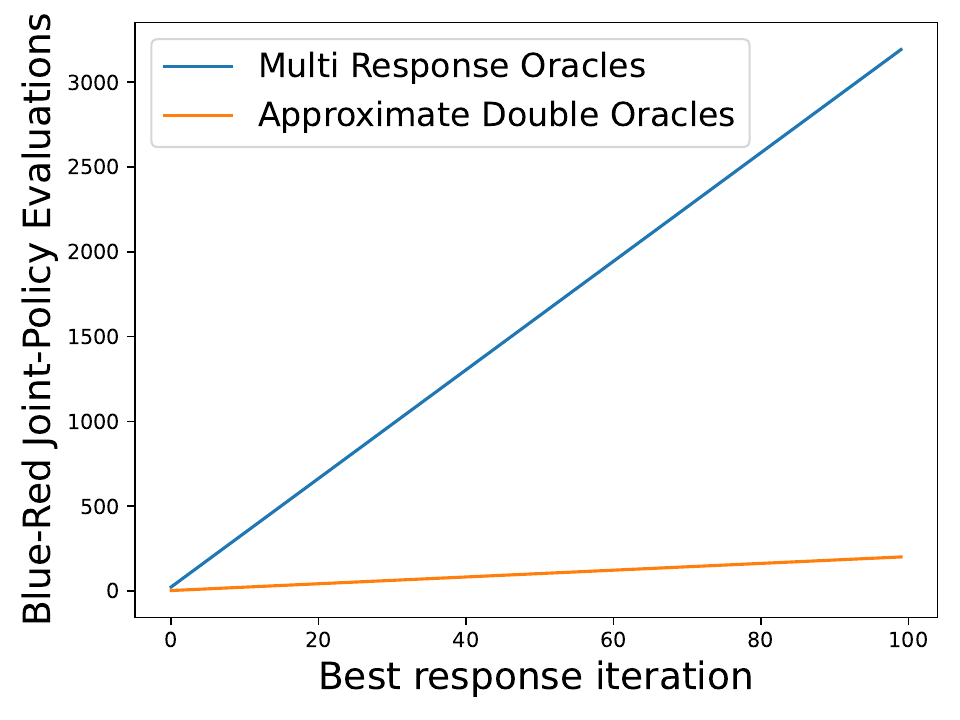}}}

\caption{In the plots above we illustrate the computational complexities of
\MRO compared to the \ADO algorithm with respect to the size of the increase in the size
of the empirical payoff matrix (\autoref{fig:payoff_table_growth}) and the
number of new cells that need to be computed in each iteration (\autoref{fig:number_of_augmentations}).}
\label{fig:mro_computaitonal_impact}
\end{figure}

\begin{figure}[H]
\centering
\includegraphics[width=0.5\columnwidth]{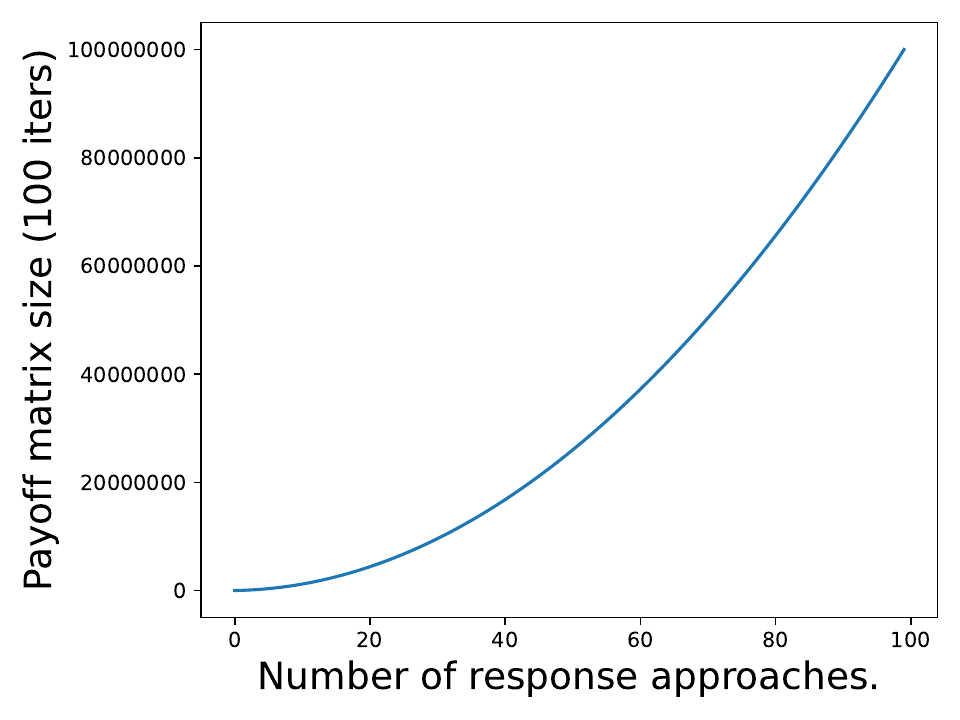}
\caption{We depict the impact of the number of methods used from computing responses
within each response iteration on the size of the payoff matrix.}
\label{fig:impact_of_response_methods}
\end{figure}

We illustrate the computational impact of using our 
proposed \MRO formulation, compared to the original \ADO algorithm.
First we highlight the exponential growth of the payoff table following each
response iteration for \MRO, compared to the quadratic growth for the \ADO algorithm, in 
\autoref{fig:payoff_table_growth}.
For this plot we assume that both \blue and \red use four approaches
for computing responses, \ie $m=n=4$.
Next, in \autoref{fig:number_of_augmentations}, 
we consider the number of additional cells that are added to the empirical
payoff matrix in each iteration, 
again assuming four response approaches for
\blue and \red. 
We observe that in iteration 100 over 3,000 new cells would need to
be added to the payoff matrix.
Finally, in \autoref{fig:impact_of_response_methods} we depict the 
exponential growth in the size of the payoff matrix after 100 ABR 
iterations when varying the number of response approaches
for \MRO.

\section{Preliminary CAGE Challenge 2 Run} \label{appendix:prelminary_cc2}

\begin{figure}[H]
\centering
\includegraphics[width=\columnwidth]{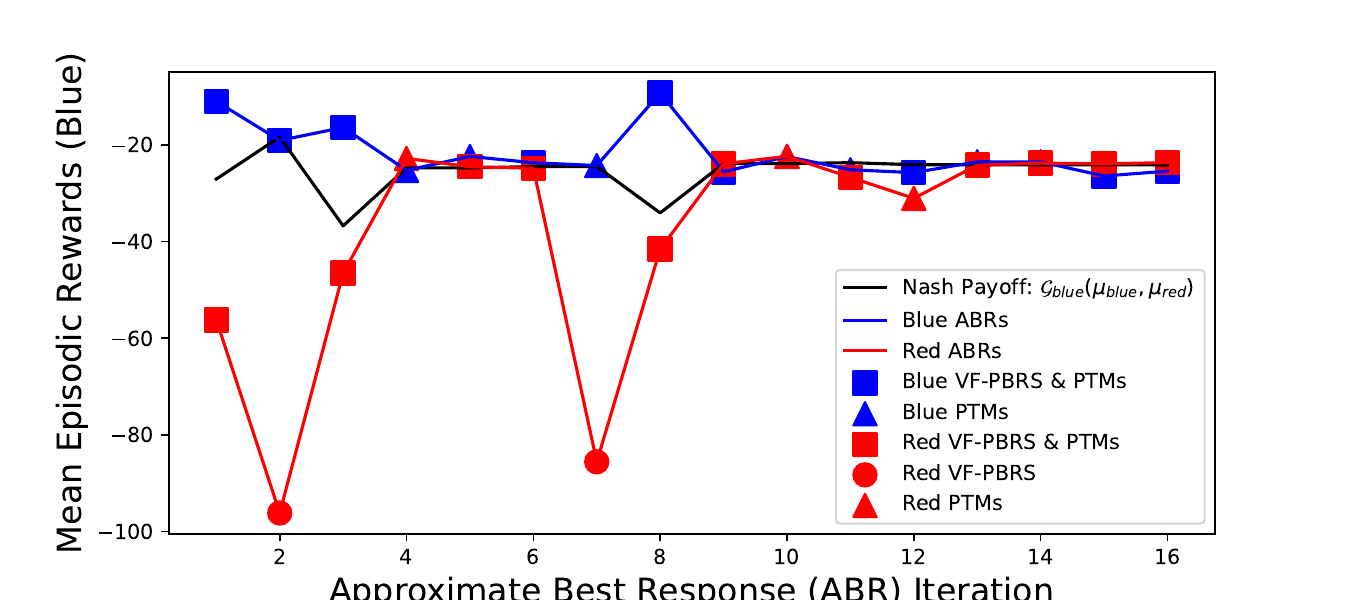}
\caption{A preliminary CAGE Challenge 2 run, comparing different
types of oracles. 
While four oracles (VF-PBRS; PTMs; VF-PBRS \& PTMs, and; Vanilla)
are used in each iteration, we only depict the best performing response.
We find the majority of ABRs were obtained by oracles using VF-PBRS \& PTMs
for both Blue and Red.
Although Red computed two noteworthy responses using only VF-PBRS in iterations
2 and 7.}
\label{fig:preliminary_cc2_run}
\end{figure}

During one of our early preliminary runs within CAGE Challenge 2 we
evaluated the impact of using: i.) VF-PBRS; ii.) initialising from PTMs;
iii.) VF-PBRS \& PTMs, and; Vanilla runs without PTMs and VF-PBRS.
Results are shown in \autoref{fig:preliminary_cc2_run}.

We found that for both Blue and Red oracles using VF-PBRS \& PTMs
returned the majority of ABRs.
Interestingly, in this preliminary run we also had two Red VF-PBRS
ABRs that converged upon significantly better responses than 
the Red mixture in iterations 2 and 7, already hinting at the
benefits of using both PTM and freshly initialised runs during training.

\section{Delayed Convergence Examples} \label{appendix:additional_evals:cc2}

\begin{figure}[H]
\centering
\includegraphics[width=\columnwidth]{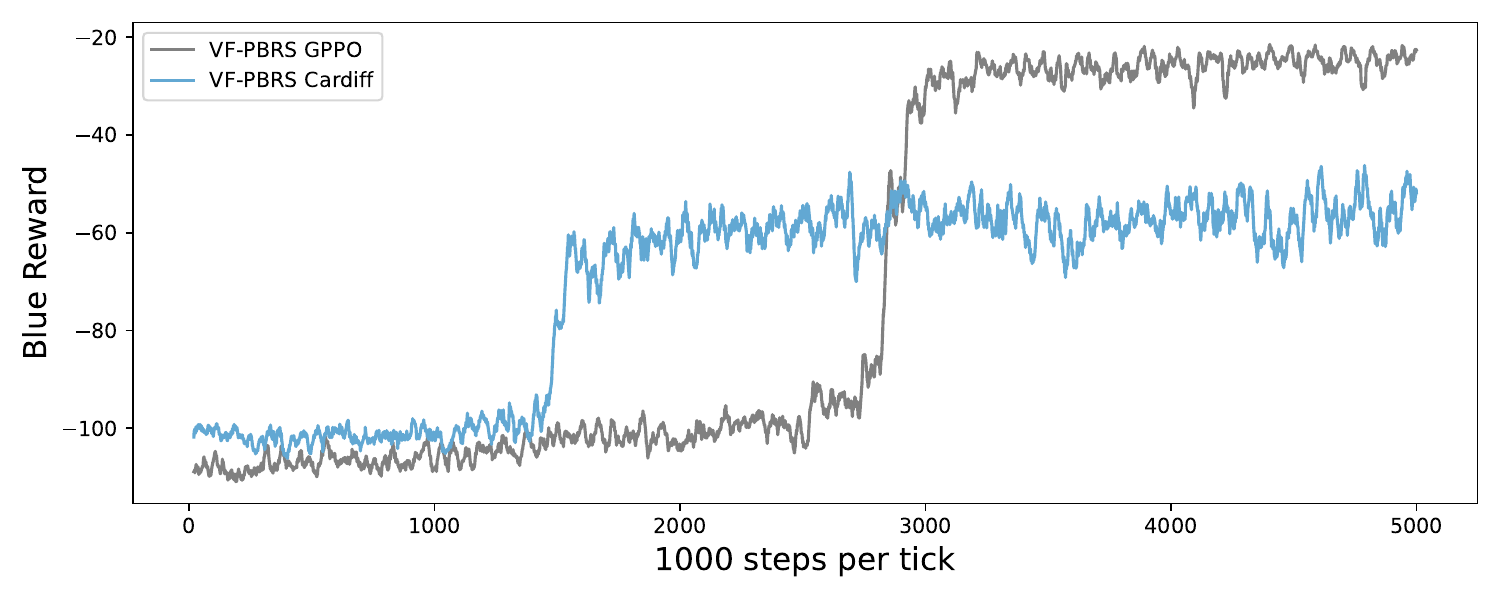}
\caption{An example of ABRs from preliminary trial
runs, which made noteworthy improvements after
several million training steps.}
\label{fig:late_bloomers}
\end{figure}

During preliminary trial runs we found that when training Blue agents (without
PTMs), they would often achieve significant improvements after a large number
of training steps.
This is illustrated in \autoref{fig:late_bloomers}, where GPPO and Cardiff ABRs
(in the third ABR iteration of a preliminary run) significantly improved their
policies after 2.5M and 1.5M time-steps respectively.
Therefore, we turn to PTMs, 
to address the long wall-times associated with these 
\say{full} runs (approximately one day per run). 

\section{Policy Characteristics -- CAGE Challenge 2} \label{appendix:additional results}

In this section we expand upon the comparison between 
the starting point of our training process, $\pi_{Blue}^o$ (the parameterisation available in the repository
maintained by \cite{CC2KEEP}), and the final mixture $\mu_{Blue}$. 
%
We note that the final mixture agent consists of GPPO agents trained 
using \cite{CC2KEEP}'s approach.
The difference between $\pi_{Blue}^o$ and the policies used by
$\mu_{Blue}$ is the opponents faced during training.

To recap, as our empirical game in \autoref{fig:cc2_payoff_matrix} illustrates,
the majority of the Blue policies with an above zero sampling weighting
within the final mixture generalise well against all the responses 
computed by Red during the MRO run.
In contrast, for $\pi_{Blue}^o$ there
exist a number of Red responses capable of forcing it
into taking costly actions.
The second Red ABR, $\pi_{Red}^2$,
has the biggest impact on $\pi_{Blue}^o$, reducing the Blue agent's payoff to
$-117.82$.
In contrast, $\mu_{Blue}$ achieves a mean payoff of $-21.96$ against $\pi_{Red}^2$.

\begin{figure}[H]
\centering
\begin{minipage}{\columnwidth}
\centering
\textbf{\footnotesize \textcolor{red}{Red Targets}}
\end{minipage}
\subfloat[Time-steps 1--5]{
\label{fig:original_blue:timesteps_1-5:aca_targets}
\includegraphics[width=0.5\columnwidth]{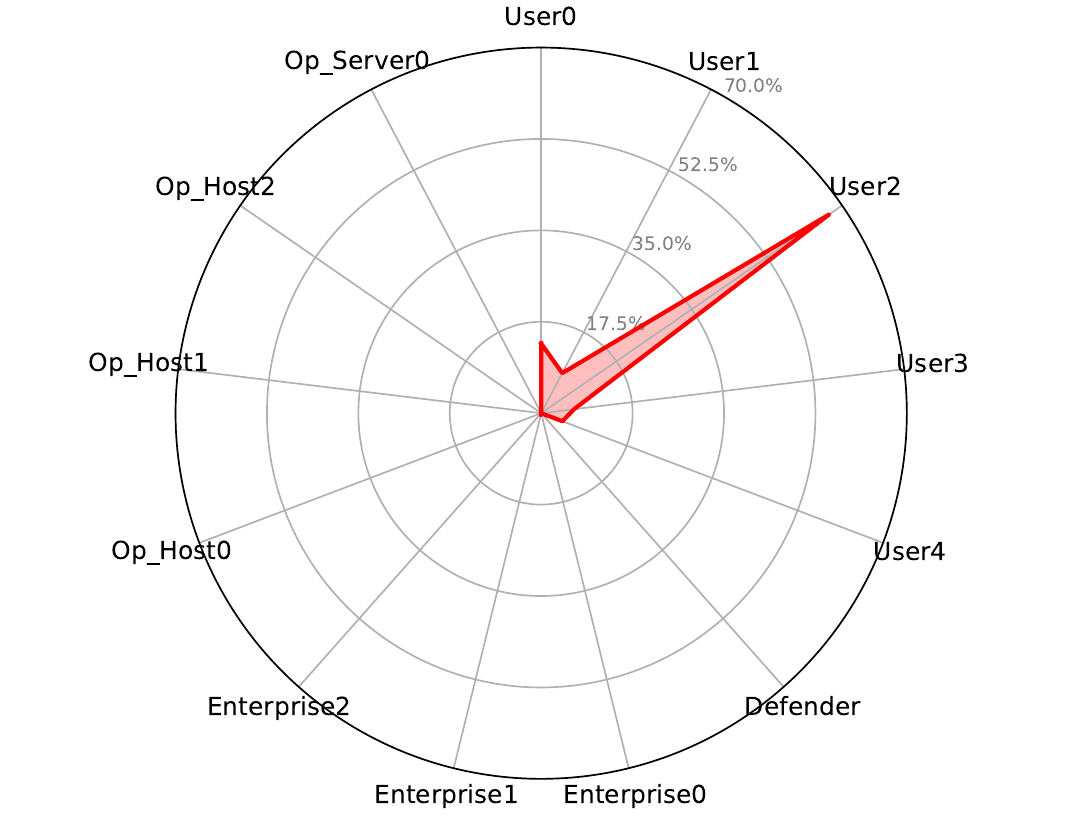}}
\subfloat[Time-steps 6--100]{
\label{fig:original_blue:timesteps_6-100:aca_targets}
\includegraphics[width=0.5\columnwidth]{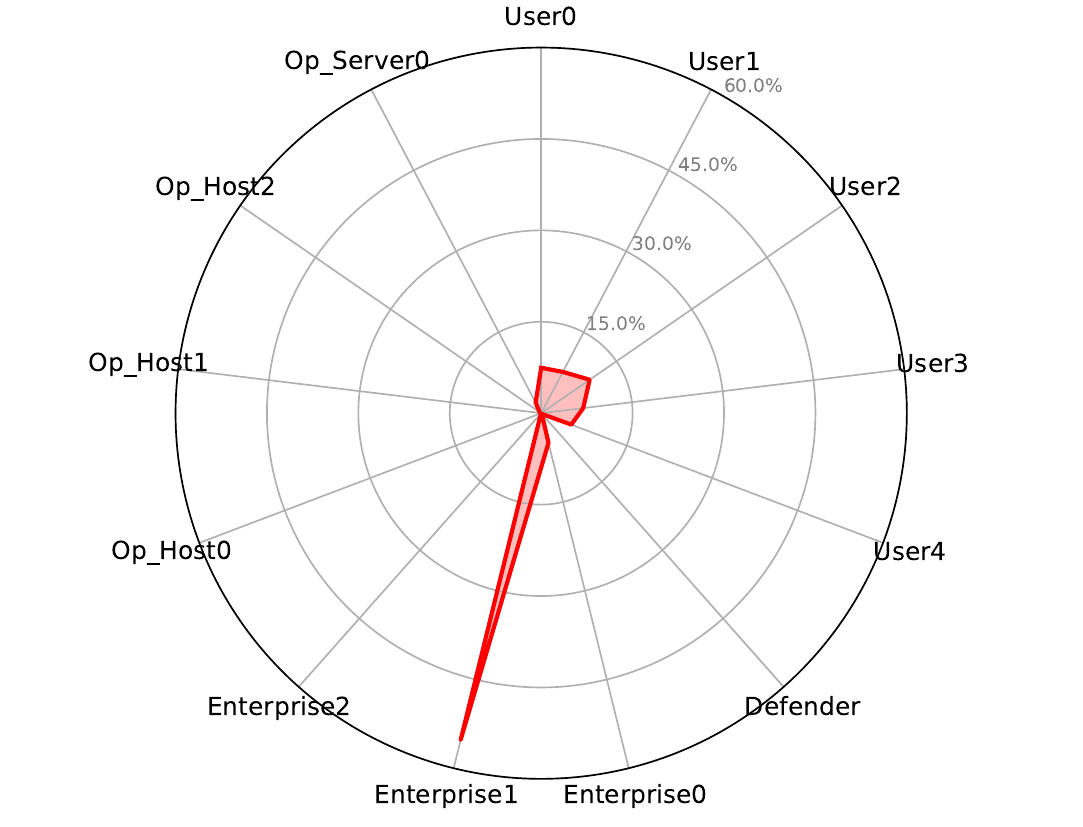}}

\begin{minipage}{\columnwidth}
\centering
\textbf{\footnotesize \textcolor{blue}{Blue Targets}}
\end{minipage}
\subfloat[Time-steps 1--5]{
\label{fig:original_blue:timesteps_1-5:acd_targets}
\includegraphics[width=0.5\columnwidth]{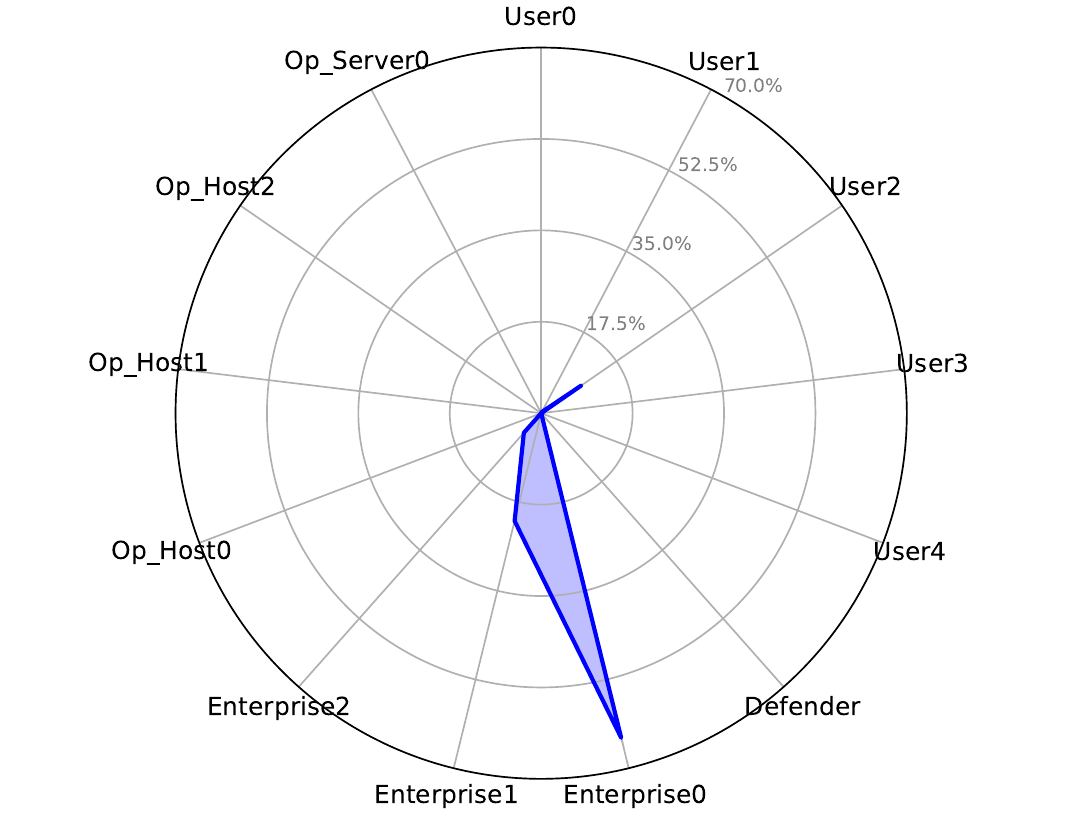}}
\subfloat[Time-steps 6--100]{
\label{fig:original_blue:timesteps_6-100:acd_targets}
\includegraphics[width=0.5\columnwidth]{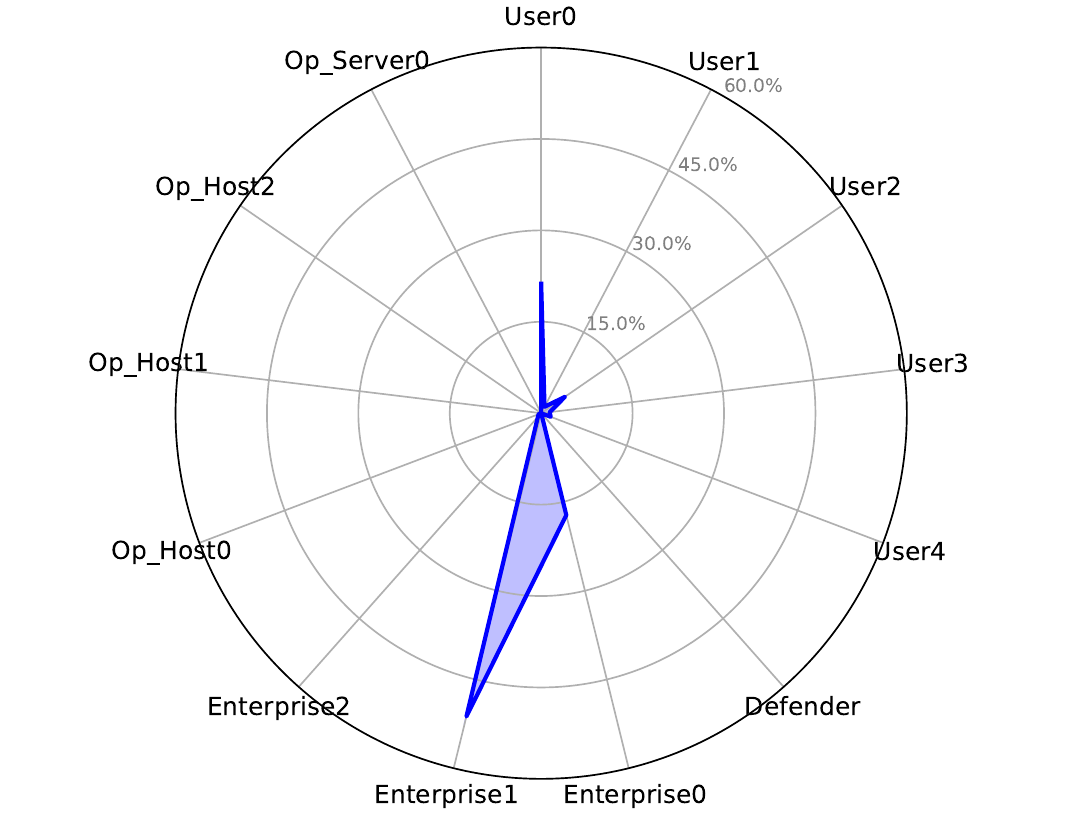}}
\caption{Nodes targeted by the original GPPO parameterisation $\pi_{Blue}^o$ and the second Red ABR $\pi_{Red}^2$ when facing each other.
Values are averaged over 100 evaluation episodes.
}
\label{fig:original_blue:RedAbr2:TargetHosts}
\end{figure}

We find that $\pi_{Red}^2$'s success against $\pi_{Blue}^o$ 
comes from initially targeting \texttt{User2} during the first 5 time-steps, and subsequently
shifting its attention to \texttt{Enterprise1} (See~\autoref{fig:original_blue:RedAbr2:TargetHosts}).
In contrast, Blue spends the first 5 steps launching decoy services on \texttt{Enterprise0}.
Red meanwhile obtains the IP address of \texttt{Enterprise1} upon gaining a hold of \texttt{User2}
(see the network digram in \autoref{fig:cc2_network_diagram}).
From this point on the two agents are locked in a cycle 
where $\pi_{Blue}^o$ restores \texttt{Enterprise1}, only for Red to regain 
access.
%

\begin{figure}[H]
\centering
\subfloat[Blue Targets]{
\label{fig:mixture_blue_vs_cherry_red:acd_targets}
\includegraphics[width=0.5\columnwidth]{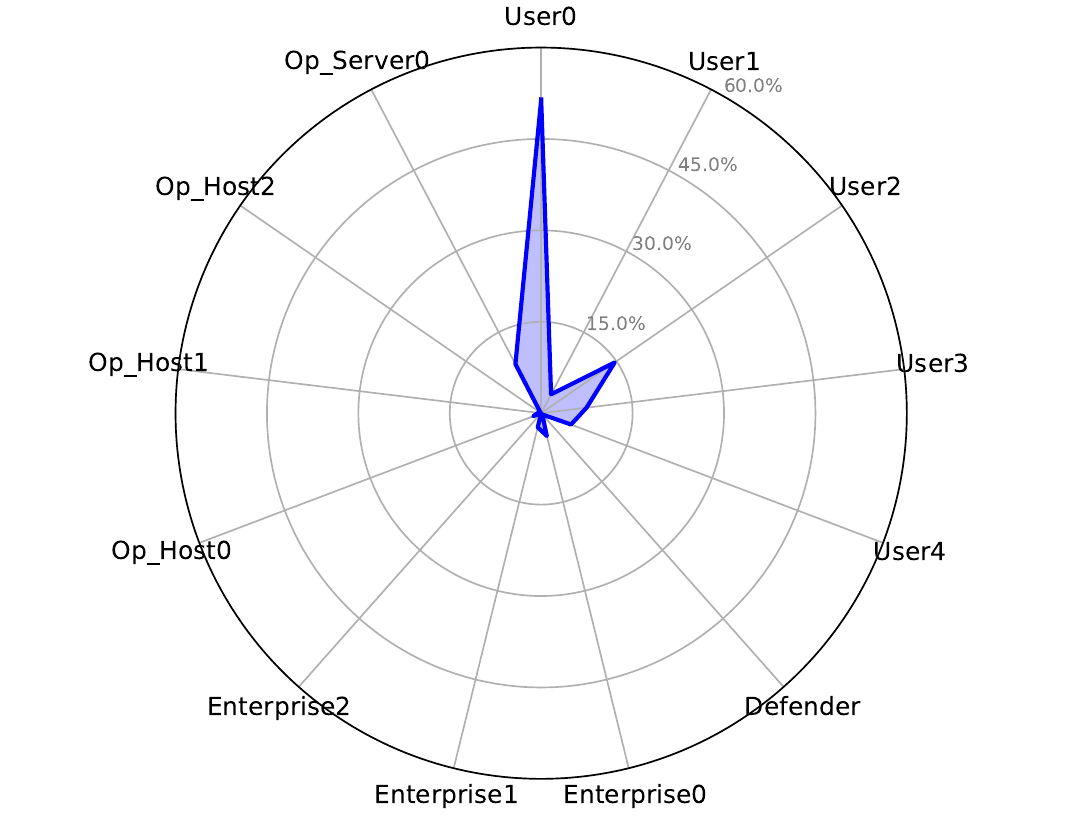}}
\subfloat[Red Targets]{
\label{fig:mixture_blue_vs_cherry_red:aca_targets}
\includegraphics[width=0.5\columnwidth]{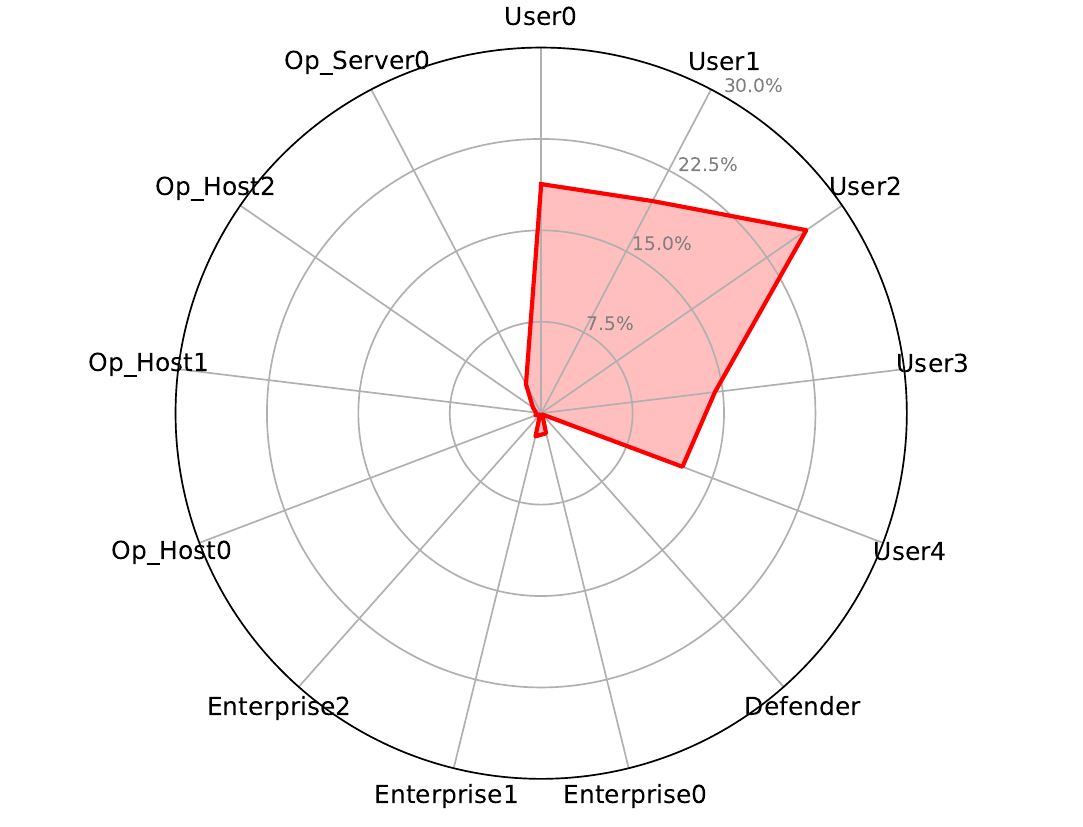}}

\caption{Policy characteristics, actions and targets from time-steps 1 - 100 for the final Blue mixture agent $\mu_{Blue}$ against $\pi_{Red}^2$.}
\label{fig:mixture_blue_vs_cherry_red}
\end{figure}

In contrast, under $\langle \mu_{Blue}, \pi_{Red}^2 \rangle$ ($\G{Blue}{\mu_{Blue}, \pi_{Red}^2}=-21.96$),
$\mu_{Blue}$ proactively analyses and restores user hosts,
resulting in $\pi_{Red}^2$ being mostly contained to the user 
network~\autoref{fig:mixture_blue_vs_cherry_red}.
In fact, the majority of $\mu_{Blue}$'s actions focus on \texttt{User0} (See \autoref{fig:mixture_blue_vs_cherry_red:acd_targets}).
This indicates that instances where Red manages to gain user or system level privileges on the user hosts 1 -- 4 are rare.
 
To complement the spider plots in Figures 
\ref{fig:original_blue:RedAbr2:TargetHosts} and 
\ref{fig:mixture_blue_vs_cherry_red},
we provide a breakdown of the action distributions for
Blue and Red agents conditioned on the level of Red access 
in Tables \ref{tab:original_blue_vs_cherry_red_privileged_access:blue}
-- 
\ref{tab:mixture_blue_vs_cherry_red_privileged_access:red}.
The tables enable a comparison of the actions taken by 
Blue and Red under the  joint policy profiles $\langle \pi_{Blue}^o, \pi_{Red}^2 \rangle$ in
and $\langle \mu_{Blue}, \pi_{Red}^2 \rangle$, respectively.
Specifically, they list the percentage of actions 
used on listed nodes under three conditions: 
i.) Actions targeting
nodes before User/Privileged access by Red has been acquired; 
ii.) Actions upon Red obtaining user access, and;
iii.) Actions upon Red obtained privileged (i.e., SYSTEM/root) access.
\begin{table}[H]
\centering
\tiny
\resizebox{1.0\columnwidth}{!}{
\begin{tabular}{|c||c|c|c|c|c|c|c|c|c|c|c|}
\hline
\multicolumn{12}{|c|}{\textbf{\textcolor{blue}{Blue Actions Overview for $\langle \pi_{Blue}^o, \pi_{Red}^2 \rangle$}}} \\
\hline 
\hline
\rotatebox[origin=c]{90}{\textbf{Host}} & 
\rotatebox[origin=c]{90}{\textbf{Analyse}} & 
\rotatebox[origin=c]{90}{\textbf{DecoyApache}} & 
\rotatebox[origin=c]{90}{\textbf{DecoyFemitter}} & 
\rotatebox[origin=c]{90}{\textbf{ DecoyHarakaSMPT }} & 
\rotatebox[origin=c]{90}{\textbf{DecoySSHD}} & 
\rotatebox[origin=c]{90}{\textbf{DecoySmss}} & 
\rotatebox[origin=c]{90}{\textbf{DecoySvchost}} & 
\rotatebox[origin=c]{90}{\textbf{DecoyTomcat}} & 
\rotatebox[origin=c]{90}{\textbf{DecoyVsftpd}} & 
\rotatebox[origin=c]{90}{\textbf{Remove}} & 
\rotatebox[origin=c]{90}{\textbf{Restore}} \\
\hline
\hline
\multicolumn{12}{|c|}{\textbf{Blue Actions before User/Privileged Access}} \\
\hline
\hline 
Enterprise0 & - & -& 0.77 & 0.06 & -& 0.02 & -& 0.12 & 0.03 & -& - \\ 
Enterprise1 & - & -& 0.13 & - & -& - & -& - & - & -& 0.87 \\ 
Enterprise2 & - & -& 0.99 & - & -& 0.01 & -& - & - & -& - \\ 
Op\_Host0 & - & -& 1.00 & - & -& - & -& - & - & -& - \\ 
Op\_Server0 & - & -& 1.00 & - & -& - & -& - & - & -& - \\ 
User1 & - & -& 0.36 & - & -& - & -& - & - & -& 0.64 \\ 
User2 & - & -& - & - & -& - & -& - & - & -& 1.00 \\ 
User3 & 0.30 & -& 0.23 & - & -& - & -& - & - & -& 0.47 \\ 
User4 & 0.11 & -& - & - & -& - & -& - & - & -& 0.89 \\ 
\hline
\hline
\multicolumn{12}{|c|}{\textbf{Blue Actions in response to User Access}} \\
\hline
\hline
Enterprise1 & - & -& 0.20 & - & -& -& -& - & -& -& 0.80 \\
User2 & - & -& 0.17 & 0.04 & -& -& -& 0.13 & -& -& 0.65 \\
User3 & 0.03 & -& 0.03 & 0.05 & -& -& -& 0.87 & -& -& 0.03 \\
User4 & - & -& 0.04 & - & -& -& -& 0.91 & -& -& 0.04 \\ 
\hline 
\hline 
\multicolumn{12}{|c|}{\textbf{Blue Actions in response to Privileged Access}} \\
\hline 
\hline
Enterprise1 & - & -& 0.08 & - & - & - & -& - & -& -& 0.91 \\
User0 & 0.41 & -& 0.42 & - & 0.08 & 0.03 & -& - & -& -& 0.05 \\
User1 & 0.09 & -& 0.45 & - & - & - & -& 0.36 & -& -& 0.09 \\
User2 & 0.02 & -& 0.22 & 0.34 & - & - & -& 0.33 & -& -& 0.09 \\
User3 & - & -& - & 0.13 & - & - & -& 0.87 & -& -& - \\
User4 & - & -& 0.30 & 0.10 & - & - & -& 0.60 & -& -& - \\ 
\hline
\end{tabular}}
\caption{CAGE Challenge 2 -- Blue action distribution on hosts and servers under the joint-policy profiles $\langle \pi_{Blue}^o, \pi_{Red}^2 \rangle$.} 
\label{tab:original_blue_vs_cherry_red_privileged_access:blue}
\end{table}

\begin{table}[H]
\tiny
\centering
\resizebox{1.0\columnwidth}{!}{
\begin{tabular}{|c||c|c|c|c|c|c|c|c|c|c|c|}
\hline
\multicolumn{12}{|c|}{\textbf{\textcolor{blue}{Blue Actions Overview for $\langle \mu_{Blue}, \pi_{Red}^2 \rangle$}}} \\
\hline 
\hline
 \rotatebox[origin=c]{90}{\textbf{Host}} & 
 \rotatebox[origin=c]{90}{\textbf{Analyse}} & 
 \rotatebox[origin=c]{90}{\textbf{DecoyApache}} & 
 \rotatebox[origin=c]{90}{\textbf{DecoyFemitter}} & 
 \rotatebox[origin=c]{90}{\textbf{ DecoyHarakaSMPT }} & 
 \rotatebox[origin=c]{90}{\textbf{DecoySSHD}} & 
 \rotatebox[origin=c]{90}{\textbf{DecoySmss}} & 
 \rotatebox[origin=c]{90}{\textbf{DecoySvchost}} & 
 \rotatebox[origin=c]{90}{\textbf{DecoyTomcat}} & 
 \rotatebox[origin=c]{90}{\textbf{DecoyVsftpd}} & 
 \rotatebox[origin=c]{90}{\textbf{Remove}} & 
 \rotatebox[origin=c]{90}{\textbf{Restore}} \\
\hline
\hline
\multicolumn{12}{|c|}{\textbf{Blue Actions before User/Privileged Access}} \\
\hline
\hline 
Defender & - & 0.52 & - & 0.39 & -& - & - & -& 0.10 & - & - \\ 
Enterprise0 & 0.70 & 0.12 & 0.03 & 0.05 & -& - & 0.06 & -& 0.02 & - & 0.02 \\ 
Enterprise1 & - & - & 0.17 & 0.71 & -& 0.02 & 0.03 & -& - & - & 0.07 \\ 
Enterprise2 & - & - & 0.67 & 0.33 & -& - & - & -& - & - & - \\ 
Op\_Host0 & 0.21 & 0.07 & - & 0.11 & -& 0.57 & - & -& 0.04 & - & - \\ 
Op\_Host1 & 0.48 & 0.10 & - & 0.31 & -& - & - & -& 0.12 & - & - \\ 
Op\_Host2 & - & 0.20 & - & 0.65 & -& - & - & -& 0.15 & - & - \\ 
Op\_Server0 & 0.47 & 0.06 & 0.01 & 0.01 & -& - & 0.34 & -& 0.01 & 0.11 & - \\ 
User1 & - & 0.02 & 0.12 & 0.09 & -& - & - & -& - & - & 0.76 \\ 
User2 & - & 0.01 & - & - & -& - & - & -& - & - & 0.99 \\ 
User3 & 0.19 & 0.31 & - & 0.26 & -& - & - & -& 0.01 & - & 0.23 \\ 
User4 & 0.22 & 0.09 & - & 0.15 & -& - & - & -& 0.02 & - & 0.52 \\ 
\hline
\hline
\multicolumn{12}{|c|}{\textbf{Blue Actions in response to User Access}} \\
\hline
\hline
Enterprise0 & 0.98 & -& - & - & - & -& -& -& - & -& 0.02 \\ 
User1 & - & -& - & - & - & -& -& -& - & -& 1.00 \\ 
User2 & 0.68 & -& 0.04 & 0.10 & - & -& -& -& 0.01 & -& 0.18 \\ 
User3 & 0.34 & -& 0.01 & 0.63 & 0.01 & -& -& -& - & -& 0.02 \\ 
User4 & 0.74 & -& 0.03 & 0.20 & - & -& -& -& - & -& 0.03 \\ 
\hline 
\hline 
\multicolumn{12}{|c|}{\textbf{Blue Actions in response to Privileged Access}} \\
\hline 
\hline
Enterprise1 & - & - & - & - & - & -& - & -& -& - & 1.00 \\ 
User0 & 0.40 & - & 0.08 & 0.02 & 0.49 & -& 0.01 & -& -& 0.01 & - \\ 
User1 & 0.56 & - & 0.01 & 0.40 & - & -& - & -& -& - & 0.02 \\ 
User2 & 0.72 & - & - & 0.02 & - & -& - & -& -& - & 0.25 \\ 
User3 & 0.88 & - & - & 0.11 & - & -& - & -& -& - & 0.01 \\ 
User4 & 0.80 & 0.02 & 0.04 & 0.13 & - & -& - & -& -& - & 0.02 \\ 
\hline
\end{tabular}}
\caption{CAGE Challenge 2 -- Blue action distribution on hosts and servers under the joint-policy profiles $\langle \mu_{Blue}, \pi_{Red}^2 \rangle$.} 
\label{tab:mixture_blue_vs_cherry_red_privileged_access:blue}
\end{table}

Tables 
\ref{tab:original_blue_vs_cherry_red_privileged_access:blue}
and 
\ref{tab:mixture_blue_vs_cherry_red_privileged_access:blue}
show that, in contrast to $\pi_{Blue}^o$, $\mu_{Blue}$ dedicates time towards 
analysing the user hosts and restoring them when necessary.  
The policies also differ in their approach towards protecting the operational subnet.
The mixture agent $\mu_{Blue}$ is proactive, 
analysing and safe-guarding the operational 
subnet on $5.66\%$ of time-steps, despite $\pi_{Red}^2$ not posing a threat.
In contrast, $\pi_{Blue}^o$ dedicates only $0.02\%$ of time-steps
towards 
the operational subnet.

\begin{table}[H]
\tiny
\resizebox{1.0\columnwidth}{!}{
\begin{tabular}{|c||c|c|c|c|c|c|c|c|c|c|c|c|}
\hline
\multicolumn{13}{|c|}{\textbf{\textcolor{red}{Red Actions Overview for $\langle \pi_{Blue}^o, \pi_{Red}^2 \rangle$}}} \\
\hline 
 \rotatebox[origin=c]{90}{\textbf{Host }} & 
 \rotatebox[origin=c]{90}{\textbf{BlueKeep}} & 
 \rotatebox[origin=c]{90}{\textbf{ DiscoverNetworkServices }} & 
 \rotatebox[origin=c]{90}{\textbf{EternalBlue}} & 
 \rotatebox[origin=c]{90}{\textbf{ExploitRemoteService}} & 
 \rotatebox[origin=c]{90}{\textbf{FTPDirectoryTraversal}} & 
 \rotatebox[origin=c]{90}{\textbf{HTTPRFI}} & 
 \rotatebox[origin=c]{90}{\textbf{HTTPSRFI}} & 
 \rotatebox[origin=c]{90}{\textbf{HarakaRCE}} & 
 \rotatebox[origin=c]{90}{\textbf{Impact}} & 
 \rotatebox[origin=c]{90}{\textbf{PrivilegeEscalate}} & 
 \rotatebox[origin=c]{90}{\textbf{SQLInjection}} & 
 \rotatebox[origin=c]{90}{\textbf{SSHBruteForce}} \\
\hline
\hline 
\multicolumn{13}{|c|}{\textbf{Red Actions before obtaining User Access}} \\
\hline 
\hline 
Enterprise1 & - & 0.02 & 0.80 & 0.01 & 0.09 & - & - & 0.04 & - & - & 0.04 & - \\ 
User1 & 0.71 & - & 0.23 & - & - & 0.01 & 0.02 & 0.01 & - & - & 0.01 & - \\ 
User2 & - & 0.18 & - & 0.05 & 0.47 & 0.03 & 0.02 & 0.02 & 0.03 & 0.15 & 0.03 & 0.03 \\ 
User3 & 0.65 & 0.10 & 0.19 & 0.01 & 0.03 & - & - & - & - & - & 0.02 & 0.01 \\ 
User4 & 0.86 & 0.02 & 0.08 & 0.01 & 0.04 & - & - & - & - & - & - & - \\
\hline 
\hline 
\multicolumn{13}{|c|}{\textbf{Red Actions upon obtaining User Access}} \\
\hline 
\hline 
Enterprise1 & 0.29 & - & - & -& - & - & 0.43 & -& -& -& - & 0.29 \\ 
User1 & - & - & 0.10 & -& - & - & - & -& -& -& - & 0.90 \\ 
User2 & 0.96 & - & 0.03 & -& - & - & - & -& -& -& - & - \\ 
User3 & 0.11 & 0.04 & 0.04 & -& 0.07 & 0.09 & 0.61 & -& -& -& - & 0.04 \\ 
User4 & 0.12 & - & - & -& - & 0.31 & 0.23 & -& -& -& 0.04 & 0.31 \\ 
\hline
\hline 
\multicolumn{13}{|c|}{\textbf{Red Actions upon obtaining Privileged  Access}} \\
\hline 
\hline
Enterprise1 & 1.00 & - & - & - & - & - & - & - & - & - & - & -\\ 
User0 & 0.58 & - & 0.26 & 0.01 & 0.04 & 0.01 & 0.01 & 0.02 & 0.01 & 0.03 & 0.02 & -\\ 
User1 & 0.16 & - & 0.18 & - & 0.63 & - & - & - & - & - & 0.02 & -\\ 
User2 & 0.29 & 0.01 & 0.46 & 0.01 & 0.03 & 0.01 & 0.01 & - & - & 0.16 & 0.01 & -\\ 
User3 & 0.44 & - & 0.07 & 0.04 & 0.04 & - & 0.04 & 0.37 & - & - & - & -\\ 
User4 & 0.36 & 0.04 & - & - & 0.04 & - & 0.04 & 0.20 & - & - & 0.32 & -\\ 
\hline
\hline 
\end{tabular}}
\caption{CAGE Challenge 2 -- Red action distribution on hosts and servers under the joint-policy profiles $\langle \pi_{Blue}^o, \pi_{Red}^2 \rangle$.} 
\label{tab:original_blue_vs_cherry_red_privileged_access:red}
\end{table}

With respect to Red attacks against $\pi_{Blue}^o$, 
upon obtaining \texttt{Enterprise1}'s
IP address through gaining privileged access on \texttt{User2},
$\pi_{Red}^2$ begins launching \texttt{EternalBlue} attacks
against \texttt{Enterprise1} (See \autoref{tab:original_blue_vs_cherry_red_privileged_access:red}).
Here, it is worth noting that there are classes of attacks that
allow Red to jump straight to privileged access on windows machines,
including \texttt{BlueKeep} and \texttt{EternalBlue}\footnote{See action tutorials in~\cite{cage_challenge_2_announcement}.}.
Hosts on the user and enterprise subnets belong to this category.
As a result Blue $\pi_{Blue}^o$ is forced into taking the restore action
(\autoref{tab:original_blue_vs_cherry_red_privileged_access:blue}). 

\begin{table}[h]
\tiny
\centering
\resizebox{1.0\columnwidth}{!}{
\begin{tabular}{|c||c|c|c|c|c|c|c|c|c|c|c|c|}
\hline
\multicolumn{13}{|c|}{\textbf{\textcolor{red}{Red Actions Overview for $\langle \mu_{Blue}, \pi_{Red}^2 \rangle$}}} \\
\hline 
  \rotatebox[origin=c]{90}{\textbf{Host }} & 
  \rotatebox[origin=c]{90}{\textbf{BlueKeep}} & 
  \rotatebox[origin=c]{90}{\textbf{ DiscoverNetworkServices }} & 
  \rotatebox[origin=c]{90}{\textbf{EternalBlue}} & 
  \rotatebox[origin=c]{90}{\textbf{ExploitRemoteService}} & 
  \rotatebox[origin=c]{90}{\textbf{FTPDirectoryTraversal}} & 
  \rotatebox[origin=c]{90}{\textbf{HTTPRFI}} & 
  \rotatebox[origin=c]{90}{\textbf{HTTPSRFI}} & 
  \rotatebox[origin=c]{90}{\textbf{HarakaRCE}} & 
  \rotatebox[origin=c]{90}{\textbf{Impact}} & 
  \rotatebox[origin=c]{90}{\textbf{PrivilegeEscalate}} & 
  \rotatebox[origin=c]{90}{\textbf{SQLInjection}} & 
  \rotatebox[origin=c]{90}{\textbf{SSHBruteForce}} \\
\hline
\hline 
\multicolumn{13}{|c|}{\textbf{Red Actions before obtaining User Access}} \\
\hline 
\hline 
Enterprise1 & - & - & 0.04 & - & - & 0.19 & 0.65 & - & - & - & 0.12 & - \\ 
User1 & 0.62 & - & 0.18 & - & - & 0.09 & 0.03 & 0.05 & - & - & 0.02 & - \\ 
User2 & - & 0.01 & - & 0.02 & 0.17 & 0.05 & 0.04 & 0.06 & 0.01 & 0.60 & 0.03 & 0.01 \\ 
User3 & 0.54 & 0.04 & 0.23 & - & 0.14 & - & - & - & - & - & 0.03 & 0.02 \\ 
User4 & 0.52 & 0.01 & 0.25 & - & 0.22 & - & - & - & - & - & - & - \\ 
\hline 
\hline 
\multicolumn{13}{|c|}{\textbf{Red Actions upon obtaining User Access}} \\
\hline 
\hline 
Enterprise0 & 0.80 & 0.01 & 0.01 & 0.14 & - & - & - & - & -& -& - & 0.03 \\
User1 & - & - & 0.02 & - & 0.02 & - & - & - & -& -& - & 0.95 \\ 
User2 & 0.75 & - & 0.12 & 0.01 & 0.09 & - & - & 0.03 & -& -& - & - \\ 
User3 & 0.30 & 0.02 & 0.04 & - & 0.04 & 0.26 & 0.24 & 0.04 & -& -& 0.03 & 0.05 \\ 
User4 & 0.31 & - & 0.10 & - & 0.08 & 0.14 & 0.15 & 0.04 & -& -& 0.04 & 0.14 \\ 
\hline
\hline 
\multicolumn{13}{|c|}{\textbf{Red Actions upon obtaining Privileged  Access}} \\
\hline 
\hline
Enterprise1 & 1.00 & - & - & - & - & - & - & - & - & - & - & - \\ 
User0 & 0.56 & - & 0.18 & - & 0.13 & 0.03 & 0.03 & 0.05 & - & 0.01 & 0.01 & - \\
User1 & 0.38 & - & 0.15 & - & 0.46 & 0.01 & - & - & - & - & - & - \\
User2 & 0.12 & - & 0.81 & - & 0.05 & - & - & 0.01 & - & - & - & - \\
User3 & 0.27 & 0.01 & 0.08 & 0.01 & 0.08 & - & 0.02 & 0.53 & - & 0.01 & - & - \\
User4 & 0.22 & - & 0.01 & - & 0.07 & 0.05 & 0.01 & 0.34 & 0.02 & 0.03 & 0.23 & 0.02 \\
\hline
\hline 
\end{tabular}}
\caption{CAGE Challenge 2 -- Red action distribution on hosts and servers under the joint-policy profiles $\langle \mu_{Blue}, \pi_{Red}^2 \rangle$.} 
\label{tab:mixture_blue_vs_cherry_red_privileged_access:red}
\end{table}

\section{CAGE Challenge 4 Results} \label{appendix:additional_results_cc4}

\begin{figure}[h]
\centering
\includegraphics[width=1.0\columnwidth]{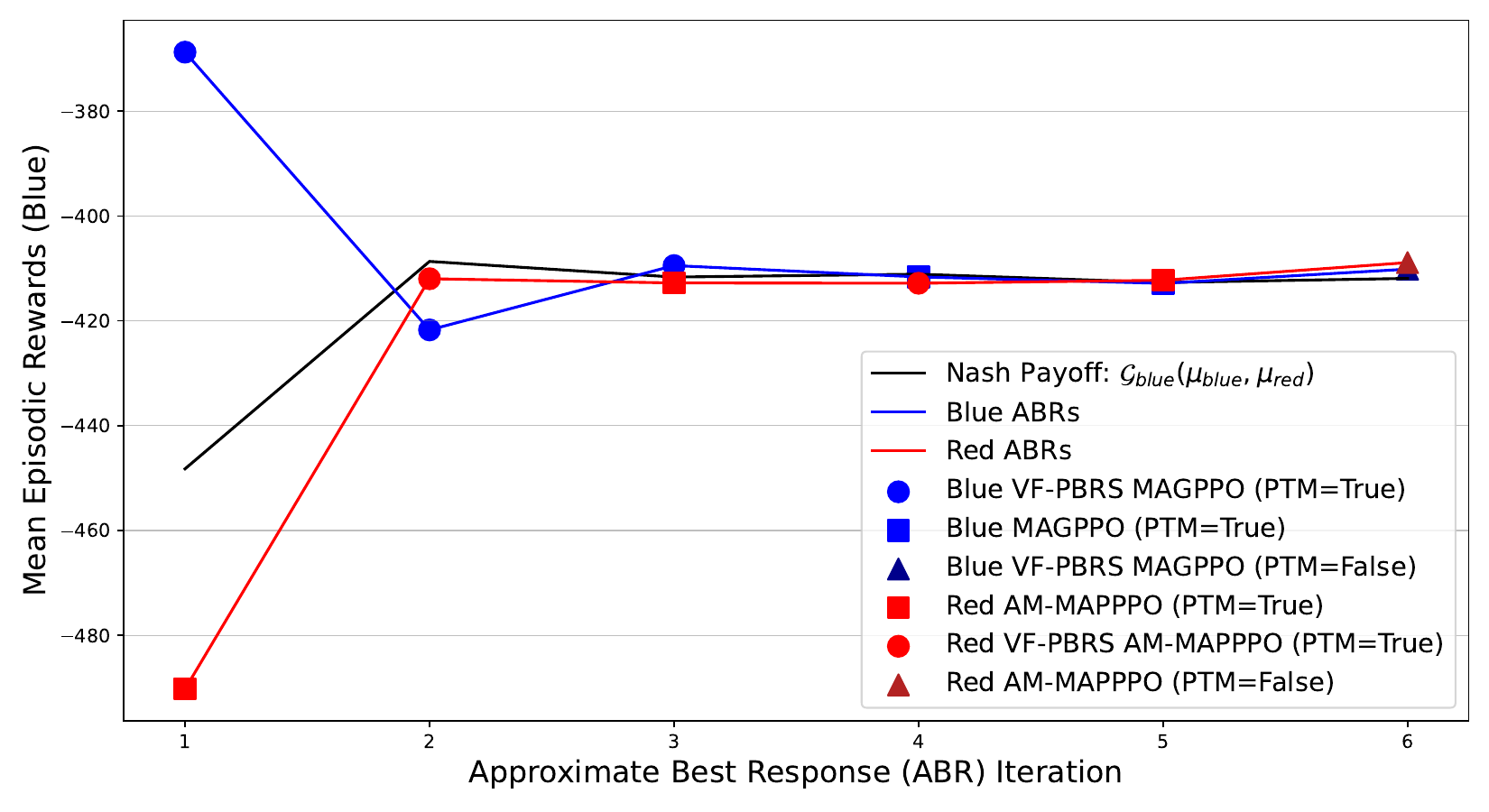}
\caption{A depiction of Blue and Red ABRs from our MRO run on CC4.
Rewards are plotted from Blue's perspective.
Both agents are unable to find ABRs that significantly improve 
on the Nash payoff (the black line) following the first ABR iteration.
}
\label{fig:cc4_ado_run:lineplot}
\end{figure}

As discussed in \autoref{sec:cc4:results}, 
upon applying MRO to CC4 we observe convergence
following only a single ABR iteration.
This is captured in \autoref{fig:cc4_ado_run:lineplot};
where following ABR iteration 1 both defending and attacking
agents are unable to significantly improve on the Nash
payoff (the black line).
With respect to the two oracles settings, 
VF-PBRS oracles compute the first three ABRs
for Blue, followed by two iterations where
Vanilla responses outperform VF-PBRS.
The final Blue ABR is also produced by a VF-PBRS oracle.

\begin{figure}[h]
\centering
\includegraphics[width=\columnwidth]{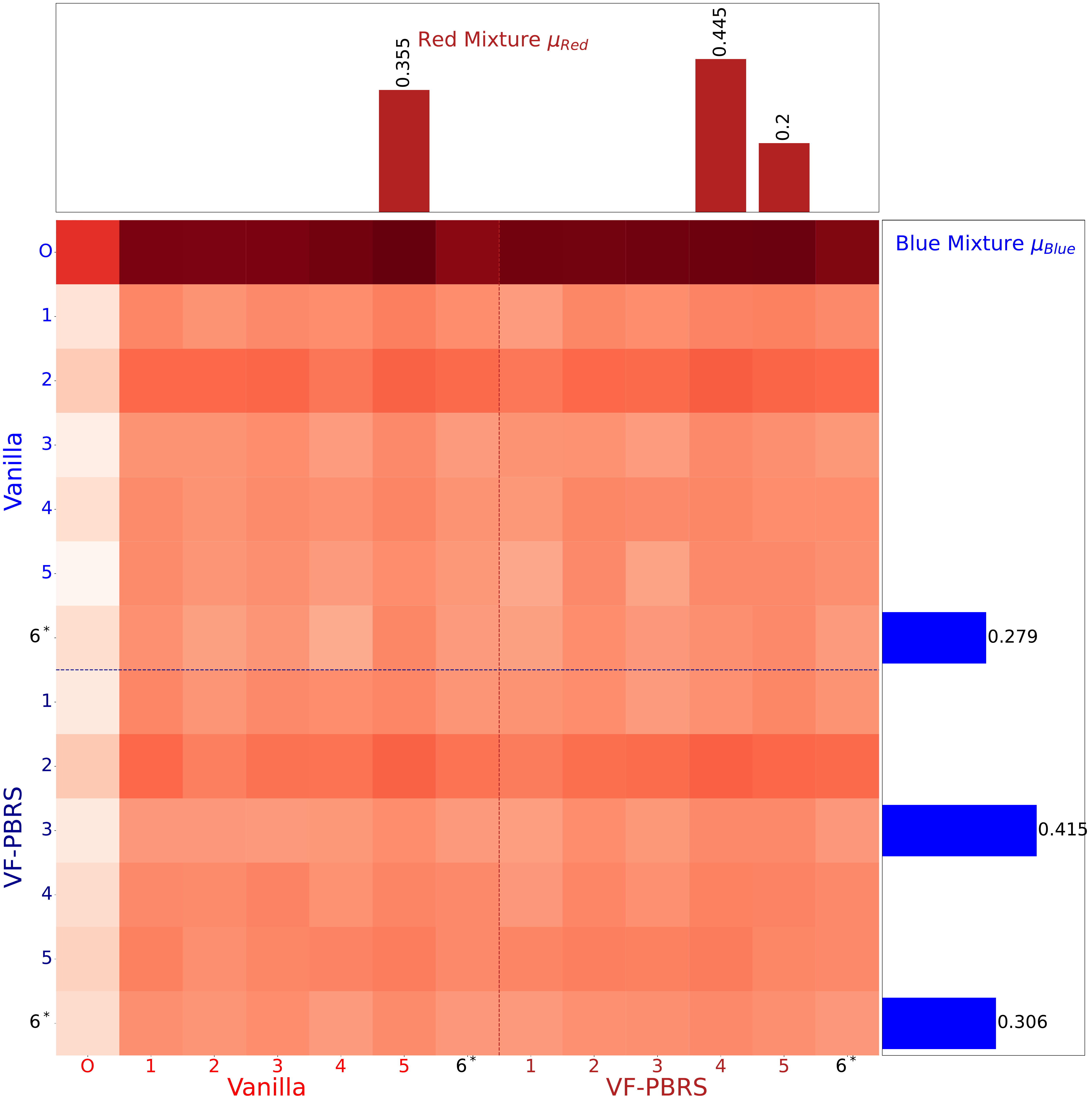}
\includegraphics[trim={0cm 0.5cm 0cm 0cm},width=\columnwidth]{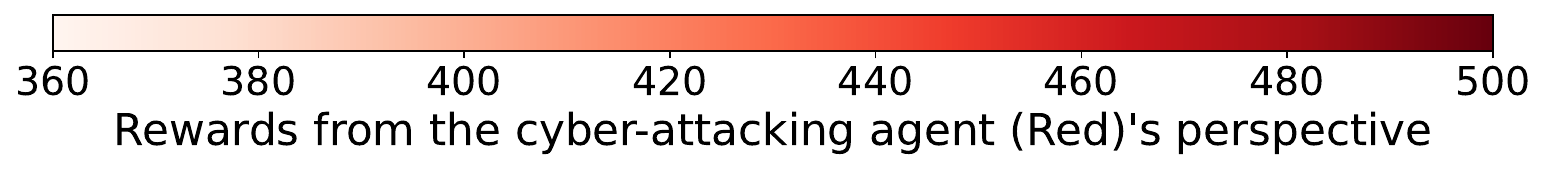}
\caption{An illustration of CC4's empirical game and Nash mixtures.
Cells represent the mean episodic reward for each policy pairing (100 evaluation episodes).
Darker cells represent match-ups that are favorable for Red.
X and Y ticks indicate the oracle used--VF-PBRS (PBRS) and Vanilla (VAN)--and 
the ABR iteration in which a response was learnt. 
The top row and left column for Blue and Red represent the 
original/starting parameterisations (O). 
Responses in iteration 6 are computed without using PTMs for
initialisation. 
These, \say{full} runs consist of 5M environment steps, 
compared to the 2.5M used when initialising from PTMs.
}
\label{fig:cc4_ado_run:heatmap}
\end{figure}

For Red only 2/6 ABRs originate from a VF-PBRS oracle.
Interestingly 2/3 responses featured in the final
Red mixture stem from VF-PBRS oracles (\autoref{fig:cc4_ado_run:heatmap}).
Similarly, the final Blue mixture also features two VF-PBRS responses,
and one vanilla response. 
Both of the final Blue responses computed without PTMs are present in the 
final mixture.  
However, upon inspecting the empirical game 
it becomes apparent that there are no significant differences between VF-PBRS and Vanilla responses 
(with the exception of VAN-1 and PBRS-4). 
The top row of the payoff matrix meanwhile shows that the original Cybermonic
parameterisation is vulnerable against all the computed Red responses.

\section{Why do we need mixed strategies?} \label{appendix:benefits_of_mixtures}

Due to our empirical games for CC2\&4  being finite zero-sum games, 
a Nash solver can be used to find the \emph{optimal}
weighting for each set of policies.
This is enabled by us selecting a zero-sum reward for training, 
with Red receiving the negation of the Blue reward.  
Therefore, Red is in effect learning to degrade Blue's performance.

\begin{figure}[h]
\centering
\subfloat[Performance of each Blue response against the final Red mixture $\mu_{\red}$.]{
\label{fig:pure_strategy_analysis_cc2_blue}
\includegraphics[width=1\textwidth]{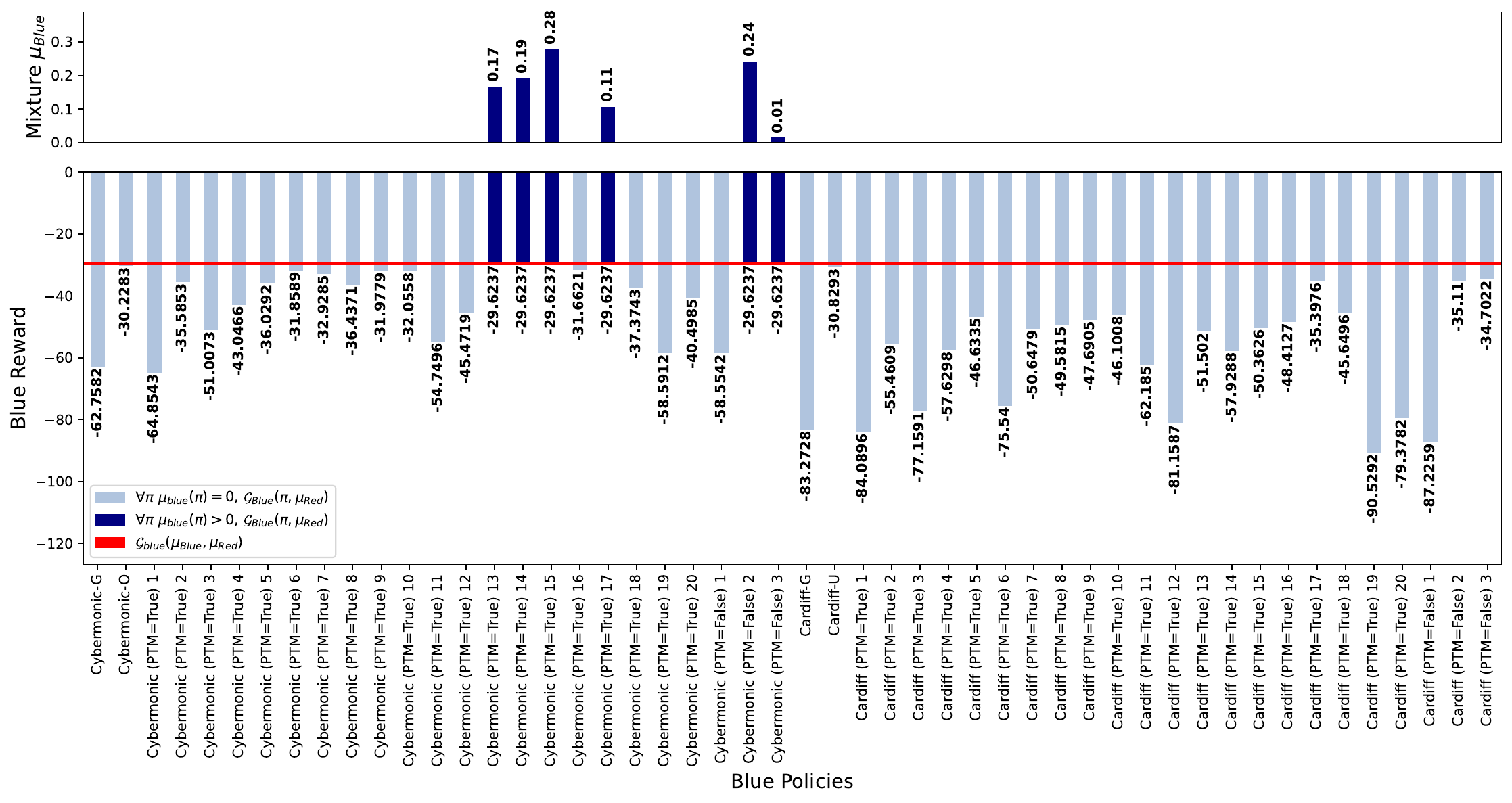}}

\subfloat[Performance of each Red response against the final Blue mixture $\mu_{\blue}$.]{
\label{fig:pure_strategy_analysis_cc2_red}
\includegraphics[width=1\textwidth]{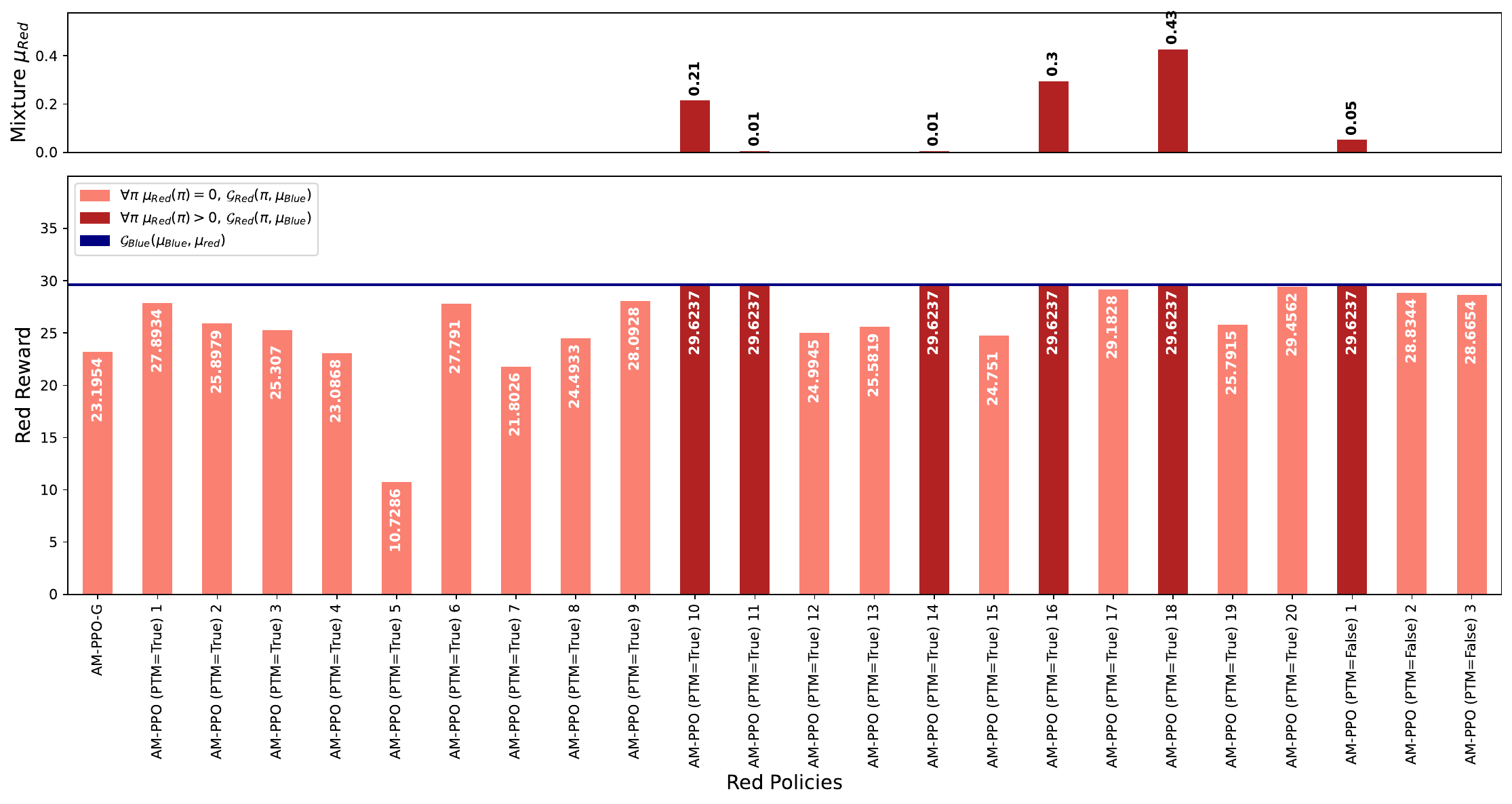}}
\caption{This figure provides a pure strategy analysis for individual \blue 
(\autoref{fig:pure_strategy_analysis_cc2_blue}) and \red (\autoref{fig:pure_strategy_analysis_cc2_red})
policies against the final \red and \blue mixture agents within CAGE Challenge 2.
Within each sub-figure, 
the upper sub-plot illustrates the sampling probabilities for the individual
policies within the respective mixture $\mu$, while the lower sub-plots depict
the empirical rewards achieved by each policy against the respective mixture.
We observe that each of the policies with an above zero sampling weighting
achieve an identical max payoff against the respective mixtures. 
This is inline with one of the key properties of a mixed strategy Nash
equilibrium, where each of the pure strategies involved in the mix must
itself be a best response against the opponent mixture~\citep{fang2021introduction}. }
\label{fig:pure_strategy_analysis_cc2}
\end{figure}

\begin{figure}[h]
\centering
\subfloat[Performance of each Blue response against the final Red mixture $\mu_{\red}$.]{
\label{fig:pure_strategy_analysis_cc4_blue}
\includegraphics[width=1\textwidth]{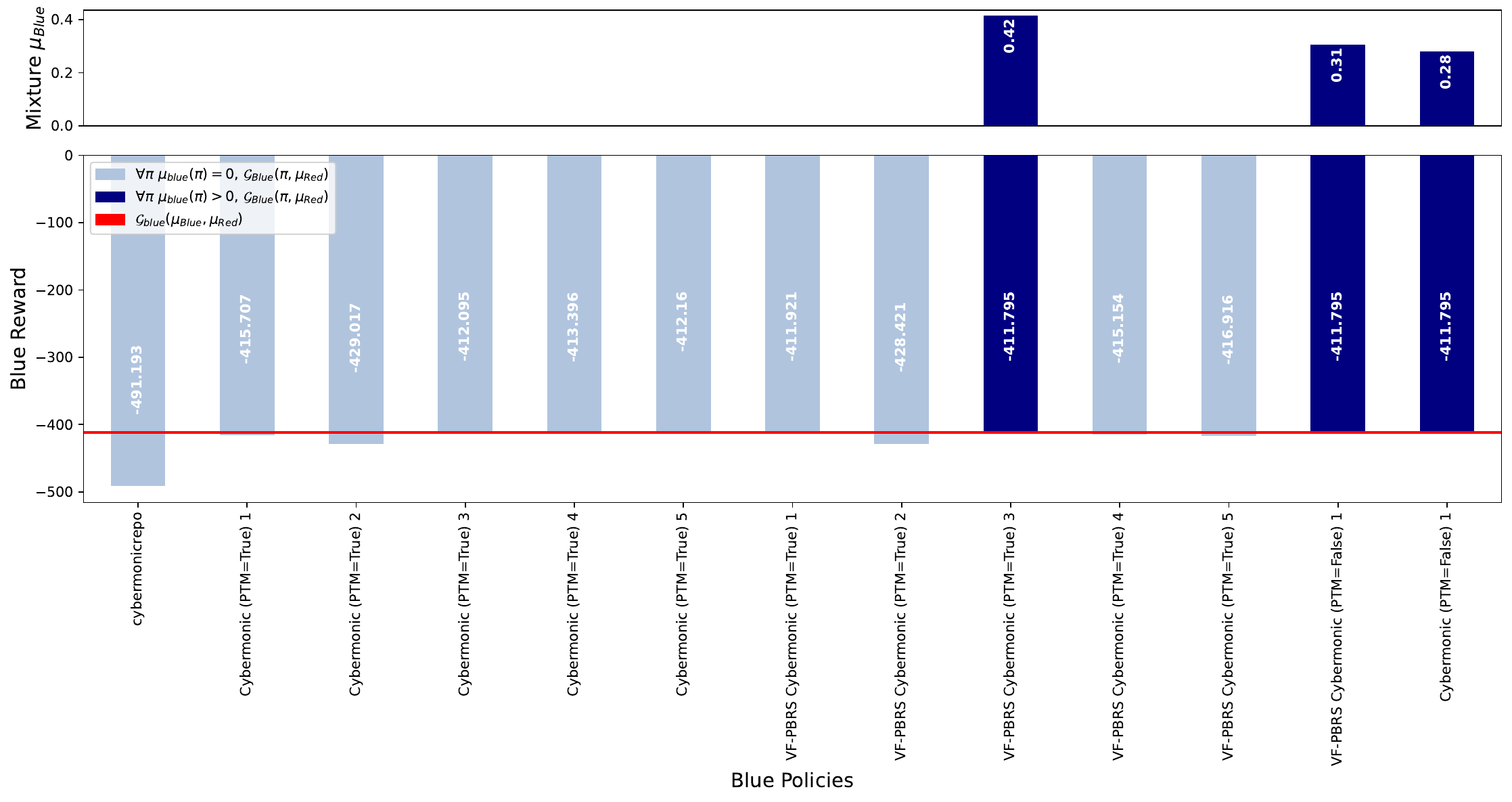}}

\subfloat[Performance of each Red response against the final Blue mixture $\mu_{\blue}$.]{
\label{fig:pure_strategy_analysis_cc4_red}
\includegraphics[width=1\textwidth]{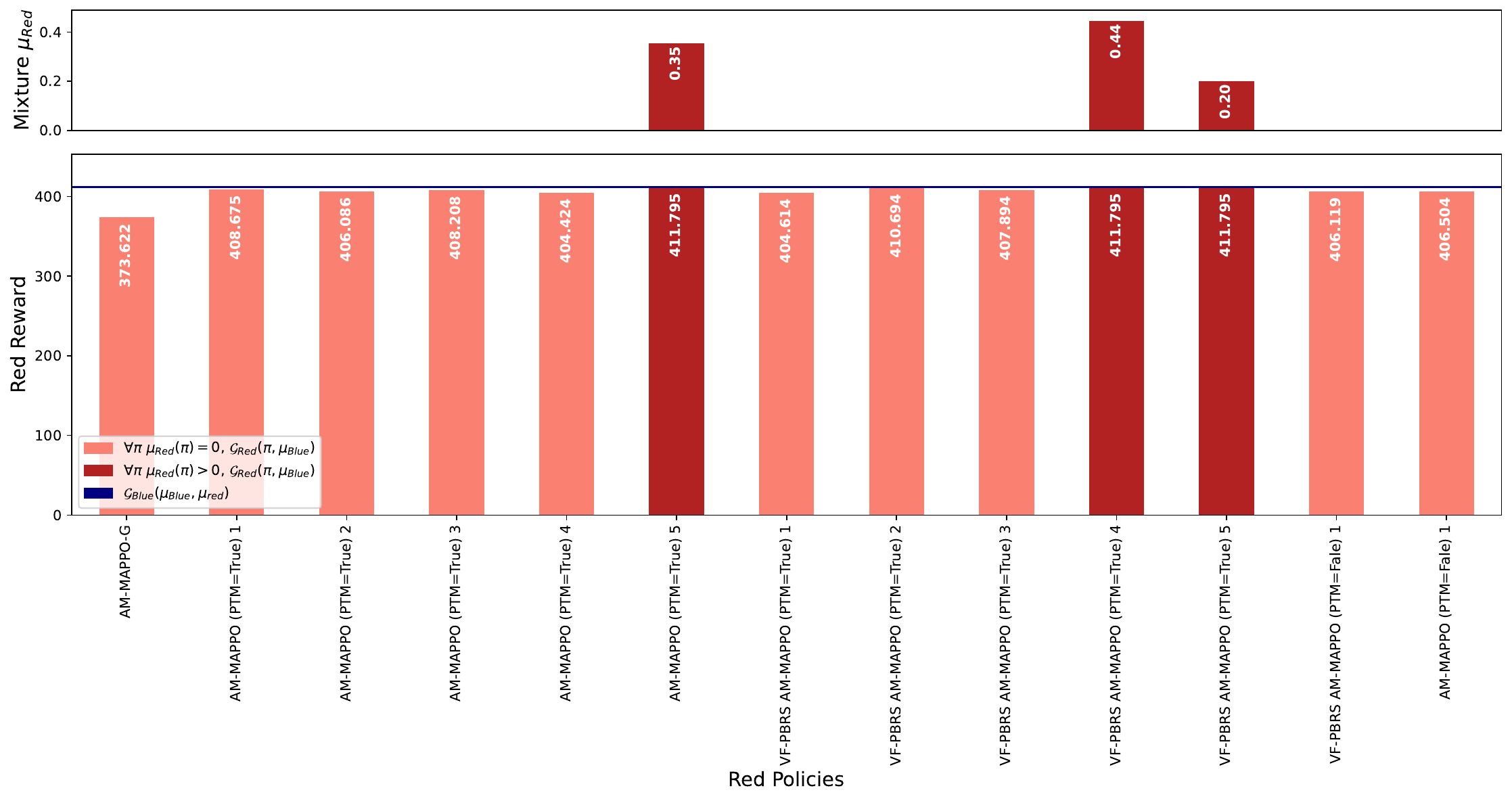}}
\caption{
As in \autoref{fig:pure_strategy_analysis_cc2},
this figure provides a pure strategy analysis for individual \blue
and \red policies, this time within CAGE Challenge 4.
In \autoref{fig:pure_strategy_analysis_cc4_blue} we observe that the 
majority of Blue responses are only marginally outperformed by the
policies within an above zero weighting in the mixture $\mu_{Blue}$.}
\label{fig:pure_strategy_analysis_cc4}
\end{figure}

At this point it is worth noting, 
each pure strategy that is found within a mixture, 
regardless of weighting, is itself a best response against the
\emph{current opponent mixture}~\citep{bjornerstedt1994nash}.
We illustrate this fact in Figures 
\ref{fig:pure_strategy_analysis_cc2} and \ref{fig:pure_strategy_analysis_cc4} for CC2 and CC4
respectively;
all policies with a mixture weighting above zero 
achieve an identical max reward against the current opponent mixture. 

Figures 
\ref{fig:pure_strategy_analysis_cc2} and \ref{fig:pure_strategy_analysis_cc4}
may give the impression that an individual best response policy could be 
deployed. 
However, this is equivalent to repeatedly using a \emph{predictable} pure strategy in the 
game of \emph{rock paper scissors}.
For example, if Blue chose to only deploy \texttt{Cybermonic (PTM=True) 14}, 
then Red could respond with selecting \texttt{AM-PPO (PTM=False) 1}, 
and achieve a payoff of $99.00$ (See \autoref{fig:cc2_payoff_matrix}).
In contrast, with Blue playing the mixture $\mu_{\blue}$, 
Red will always obtain a worse payoff upon deviating from $\mu_{\red}$
within the current empirical game.

\newpage

\onecolumn

\newpage

\newpage
\twocolumn



\end{document}